\documentclass[10pt, letter, onecolumn]{arxiv}
\usepackage[utf8]{inputenc}
\usepackage{textgreek}
\usepackage{kantlipsum, lipsum}
\usepackage{dm-colors}
\usepackage{amsmath}
\usepackage{pstricks, pst-node}
\usepackage{verbatim}
\usepackage{multirow}
\usepackage{rotating}
\usepackage{scalerel}
\usepackage{booktabs}
\usepackage{enumitem}
\usepackage{xspace}
\usepackage{comment}
\usepackage[breakable]{tcolorbox}
\usepackage{bm}
\usepackage{bbm} 
\usepackage{mathtools}
\usepackage{soul}
\usepackage{epsfig}
\usepackage{graphicx}
\usepackage{amssymb}
\usepackage{colortbl}
\usepackage{csquotes}
\usepackage{setspace}
\usepackage{colortbl}
\usepackage{bbding}
\usepackage{threeparttable}
\usepackage{tabularx,ragged2e}
\usepackage{placeins}
\usepackage{bbding}
\usepackage{algorithm}
\usepackage{algorithmic}
\usepackage{fancyvrb}
\definecolor{mainteal}{RGB}{0, 128, 128}
\definecolor{darkpastelgreen}{rgb}{0.13, 0.55, 0.13}
\definecolor{darkpastelred}{rgb}{0.55, 0.13, 0.13}
\usepackage[hang,flushmargin]{footmisc}

\usepackage{nameref}
\usepackage{varioref}
\usepackage{amssymb}%
\usepackage{lineno}
\usepackage{pifont}
\usepackage{xcolor}

\usepackage[pagebackref,breaklinks,colorlinks,
    citecolor=blue,
    filecolor=magenta,
    linkcolor=red,
    urlcolor=cyan]{hyperref}
    
\usepackage[noabbrev,capitalize]
{cleveref}
\usepackage{etoc}

\definecolor{reasoncolor}{HTML}{EE822F}
\definecolor{diagnosecolor}{HTML}{C81D31}
\definecolor{lookupcolor}{HTML}{1E5799}
\definecolor{matchcolor}{HTML}{3B82C4}
\definecolor{searchcolor}{HTML}{5BA3E0}
\graphicspath{{figures/}}
\definecolor{midgreen}{rgb}{0.0, 0.75, 0.0}
\definecolor{softblue}{HTML}{136783}

\newcommand{\ModelName}{Deep-DxSearch}

\definecolor{goodgreen}{HTML}{00B050}
\definecolor{badred}{HTML}{FF0000}

\title{\Large{End-to-End Agentic RAG System Training for \\ Traceable Diagnostic Reasoning}}

\author[1,2]{Qiaoyu Zheng} 
\author[1]{Yuze Sun}
\author[1]{Chaoyi Wu} 
\author[1,2]{Weike Zhao}
\author[1,2]{Pengcheng Qiu}
\author[1,4]{Ge Wang}
\author[3]{\\Yongguo Yu}
\author[3]{Kun Sun}
\author[5]{Jian Zhang}
\author[1]{Yanfeng Wang} 
\author[1,2,6,$\dag$]{Ya Zhang} 
\author[1,2,$\dag$]{Weidi Xie}
\affil[1]{\normalsize Shanghai Jiao Tong University, Shanghai, China \hspace{1cm}}

\affil[2]{\normalsize Shanghai AI Laboratory, Shanghai, China \authorcr \vspace{0.1cm}}

\affil[3]{\normalsize Xinhua Hospital affiliated to Shanghai Jiao Tong University School of Medicine, Shanghai, China \authorcr \vspace{0.1cm}}

\affil[4]{\normalsize Shanghai Ninth People's Hospital, Shanghai Jiao Tong University School of Medicine, Shanghai, China \authorcr \vspace{0.1cm}}

\affil[5]{\normalsize Department of Pharmaceutical and Artificial-Intelligence Sciences, Shanghai Jiao Tong University School of Medicine, Shanghai, China \authorcr \vspace{0.1cm}}

\affil[6]{\normalsize Institute of Artificial Intelligence for Medicine, School of Medicine, Shanghai Jiao Tong University, Shanghai, China}

\renewcommand{\correspondingauthor}[1]{$\dag$~Corresponding author. Email addresses: \{three-world, ya\_zhang, weidi\}@sjtu.edu.cn}

\begin{document}

\begin{abstract}
The integration of Large Language Models (LLMs) into healthcare is currently constrained by inherent knowledge limitations, hallucinations, and a fundamental disconnect from the \textbf{principles of Evidence-Based Medicine~(EBM)}. While Retrieval-Augmented Generation (RAG) offers a potential solution, current systems typically rely on static, heuristic-driven workflows that fail to capture the iterative, hypothetico-deductive reasoning characteristic of clinical experts. 
To address this, we introduce \textbf{Deep-DxSearch}, an agentic RAG system trained end-to-end via reinforcement learning~(RL) to enable traceable diagnostic reasoning. Unlike passive predictors, Deep-DxSearch operates as an active investigator, treating the LLM as an agent and a comprehensive corpus—comprising over 16,000 guideline-derived disease profiles, a structured database of 150,000+ patient records for case-based reasoning, and a massive repository of over 27 million biomedical literature-as its environment. By utilizing soft verifiable rewards that co-optimize retrieval and reasoning, the model is trained to formulate queries, evaluate evidence utility, and refine search strategies to close diagnostic gaps. 
Experiments demonstrate that our end-to-end agentic RL training framework consistently outperforms prompt-engineering and training-free RAG approaches. Across in-distribution~(ID) and out-of-distribution~(OOD) benchmarks for common and rare disease diagnosis, \ModelName{} consistently surpasses strong baselines---including GPT-4o, DeepSeek-R1, and medical-specific frameworks---achieving an average accuracy improvement of 22.7\% over the second-best model. Crucially, in a validation involving 150 real-world cases, \ModelName{} assistance elevates physicians' average diagnostic accuracy from 45.6\% to 69.1\%. These results suggest that evolving agentic systems to exploit statistical regularities in large-scale healthcare data is essential for deploying trustworthy, transparent diagnostic assistants. All data, code, and checkpoints are available at \url{https://qiaoyu-zheng.github.io/Deep-DxSearch}.


\end{abstract}

\maketitle


\section{INTRODUCTION}

The integration of Large Language Models (LLMs) into clinical workflows holds the promise of democratizing expert-level diagnostic support, potentially reducing diagnostic errors and alleviating physician burnout~\cite{singhal2023large, hager2024evaluation, qiu2025quantifying, dou2025baichuan, chen2023meditron, croxford2025evaluating, wu2023can, Asgari2025AFT, liao2025ehr}. As these models grow in capability, the prospect of an ``AI co-pilot'' that can synthesize vast amounts of patient data is becoming increasingly tangible~\cite{Liu2025AGM, zou2025rise, qiu2024llm, tu2024towards, qiu2025evolving, openai_chatgpt_health, anthropic_healthcare_2026, openevidence_home_2026}. However, the translation of these tools from research benchmarks to real-world physician assistants is currently hindered by a fundamental discordance with the \textbf{principles of Evidence-Based Medicine~(EBM)}~\cite{guyatt1992evidence, guyatt2002users, Tenny2025Evidence}. Modern clinical decision-making is not merely about generating a correct prediction; it is a rigorous process of substantiating hypotheses with established guidelines, verified literature, and historical precedents~\cite{Tenny2025Evidence}. In contrast, standard LLMs operate as ``black boxes'', generating answers via parametric intuition, a process that is prone to hallucination, and decoupled from verifiable sources~\cite{Asgari2025AFT, sandmann2025benchmark, Pham2024TowardsRM}. Consequently, clinicians are often left with a correct answer but no reliable way to verify its provenance, creating a ``trust gap'' that prevents widespread adoption in high-stakes healthcare environments. For an AI system to be clinically viable, it must transition from opaque prediction to traceable, evidence-anchored reasoning. 

To address this, \textbf{Retrieval-Augmented Generation~(RAG)} has emerged as a mechanism to ground LLMs in external data, allowing models to access up-to-date information without retraining~\cite{li-etal-2024-mmedagent,  gao2023retrieval, xiong2024benchmarking, singh2025agentic}. Yet, existing medical RAG systems remain largely static and heuristic-driven~\cite{li-etal-2024-mmedagent, wu2024medical, xia2024mmed, Xu2024BMRetrieverTL, Delile2024GraphBasedRC}, treating retrieval as a keyword-matching step rather than a reasoning process. They typically perform a ``one-shot'' retrieval based on the initial query, failing to adapt when the retrieved evidence is insufficient, irrelevant, or conflicting. This rigid approach contrasts sharply with the cognitive workflow of a human clinician, who engages in a dynamic and progressive analysis~\cite{kresevic2024optimization, leblanc2021rare, johri2025evaluation}: formulating a differential diagnosis, consulting specific guidelines to verify symptoms, searching for similar historical cases when presentations are atypical, and refining the hypothesis based on new information. When a physician encounters ambiguity, they do not stop; they dig deeper, cross-referencing sources until a consensus is reached. Existing AI frameworks lack this ``meta-cognitive'' ability to recognize their own uncertainty and actively seek out the missing information required to close the diagnostic loop.

\begin{figure}[!t]
    \centering
    \includegraphics[width=1\linewidth]{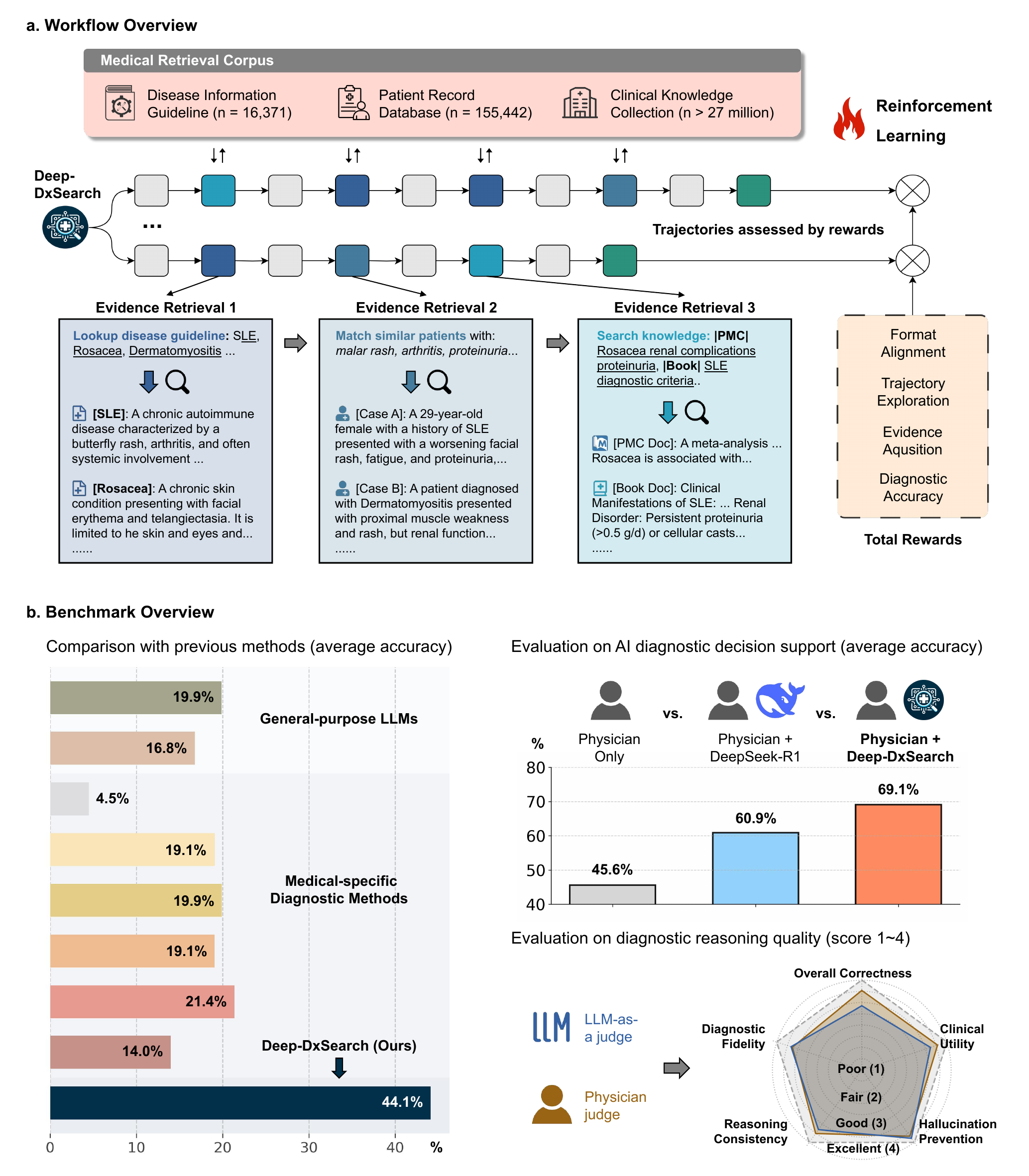}
    \caption{\textbf{Contribution Overview.}
    \textbf{a.} The proposed workflow. Top: The medical retrieval corpus serving as the search environment during both training and inference. Middle: Illustration of the \ModelName{} rollout process, where diagnostic trajectories are generated and optimized via reinforcement learning based on trajectory-level rewards. Bottom: An exemplar log demonstrating the system's traceable evidence retrieval.
    \textbf{b.} Key performance highlights across three dimensions: \ModelName{} achieves superior diagnostic accuracy compared to both general-purpose LLMs and specialized medical methods; demonstrates notable clinical utility in physician assistance, surpassing the performance of DeepSeek-R1; and consistently attains high reasoning quality (rated $>$``Good'') across five dimensions in both ``LLM-as-a-judge'' and human evaluations.}
    \label{fig:WorkOverview}
\end{figure}

Here, we introduce \textbf{Deep-DxSearch}, an agentic diagnostic system~(Fig.~\ref{fig:WorkOverview}a) that bridges this gap by learning to mimic the iterative information-seeking behavior of physicians. Unlike previous frameworks that rely on fixed prompt engineering or static retrieval pipelines, Deep-DxSearch is trained end-to-end~(optimizing under unified policy) with reinforcement learning (RL) to navigate a complex information landscape. By framing the LLM as an autonomous agent and the medical corpus as its environment, we optimize a policy that learns when to reason internally, when to query external databases, and how to synthesize heterogeneous evidence into a coherent diagnostic argument. This approach moves beyond simple question-answering; the agent is taught to formulate specific search queries, evaluate the utility of the returned documents, and iteratively refine its search strategy if the initial results are unsatisfactory. 
This mimics the ``hypothetico-deductive'' reasoning model used in clinical education, transforming the AI from a passive predictor into an active investigator.

Central to our approach is the construction of a comprehensive medical retrieval environment, designed to replicate the diverse resources available to a practicing doctor. We integrate over 16,000 guideline-derived disease profiles, a structured database of 150,000+ patient records for case-based reasoning, and a massive repository of over 27 million biomedical literature to cover rare and emerging conditions. Within this environment, Deep-DxSearch is incentivized not just for diagnostic accuracy, but for the validity of its retrieval trajectory—rewarding the agent for uncovering high-fidelity evidence that explicitly supports its conclusions. This multi-modal retrieval capability allows the system to triangulate answers, validating a diagnosis against clinical guidelines while simultaneously checking for similar historical patient presentations. By grounding the reward signal in evidence retrieval, we ensure that the model’s ``thought process'' is aligned with the availability of concrete medical facts rather than the model's likelihoods alone.

We demonstrate that this agentic reinforcement learning paradigm significantly enhances diagnostic performance~(Fig.~\ref{fig:WorkOverview}b), achieving state-of-the-art accuracy across a multi-center cohort cases covering both common and rare diseases. In a thorough evaluation spanning 8 clinical centers, Deep-DxSearch consistently outperformed both strong reasoning LLMs~\cite{hurst2024gpt, guo2025deepseek,Chen2023MEDITRON70BSM, sellergren2025medgemma} and training-free RAG or agentic baselines~\cite{xiong2024benchmarking, chen2025enhancing, chen2025cod, feng2025doctoragent}. 
Specifically, our agentic strategy surpassed standard RAG approaches by margins of 29.7\%~(p$<$0.01) on in-distribution (ID) data and 9.7\% (p$<$0.05) on out-of-distribution (OOD) data, while improving upon specialized medical foundation models by up to 31.8\%~(p$<$0.001) for rare diseases. Ablation studies reveal that these gains are driven by our process-based reward design, which simultaneously optimizes retrieval and reasoning policies—yielding a 13.7\%~(p$<$0.01) accuracy boost over target-only supervision. Beyond quantitative metrics, interpretability analyses quantify how the agent evolves during training, learning to execute more diverse search trajectories that prioritize differential diagnosis and irrelevance exclusion. Crucially, in physician-in-the-loop studies, clinicians favored Deep-DxSearch's auditable ``Chain of Evidence'' over the opaque ``Chain of Thought'' of strong reasoners like DeepSeek-R1; while the latter excel at internal logic, Deep-DxSearch provides a verifiable trail of medical literature for every claim, offering a blueprint for the next generation of trustworthy, transparent diagnostic assistants.

\section{RESULTS}

\subsection{Deep-DxSearch Enables Traceable Diagnostic Reasoning via Agentic RL}
\label{sec:framework_results}

We present \textbf{\ModelName{}}, an agentic framework that transforms diagnosis from a static classification task into a dynamic, multi-step inquiry. Unlike conventional Retrieval-Augmented Generation (RAG) systems that rely on ``one-shot'' retrieval, \ModelName{} operates as an autonomous agent orchestrating five clinical primitives: \texttt{<reason>} for hypothesis generation, \texttt{<lookup>} for guideline verification, \texttt{<match>} for case-based reasoning, \texttt{<search>} for literature review, and \texttt{<diagnose>} for final assessment.

This architecture is underpinned by a tripartite reinforcement learning (RL) framework, optimizing a policy that balances diagnostic accuracy with valid evidence retrieval and diverse trajectory exploration, the system learns to mimic the hypothetic-deductive reasoning of human clinicians. To support this, we constructed a massive medical retrieval environment comprising structured disease guidelines, validated patient records, and a semantic knowledge base large-scale biomedical documents~(see Sec.~\ref{sec:data_description} for the details of corpus construction).

In the following sections, we train and evaluate \ModelName{} against state-of-the-art foundation models and specialized diagnostic systems across a multi-center cohort of over 24k patient cases. We further present a physician-in-the-loop evaluation to validate clinical utility and provide an analysis on how end-to-end RL reshapes diagnostic policy.

\subsection{Baselines}
\label{sec:baseline}

We benchmark the performance of \ModelName{} against diverse models. First, we detail baseline LLM enhancement techniques, including both prompting and training strategies, to isolate the specific contributions of our agentic RL approach. Subsequently, we compare our whole final system against seven distinct categories of state-of-the-art diagnosis methods.

\noindent \textbf{Baseline LLM enhancement techniques.}
To rigorously evaluate the efficacy of our agentic RL approach, we compare it against four representative LLM enhancement strategies:
\textbf{(i) vanilla model with direct prompting}. 
By properly designing instructions, we can prompt the base model to preform diagnosis directly from its internal knowledge without any post-training or external retrieval, which is the most basic and widely-used method for clinical LLM alignment~\cite{wang2024prompt, ding2025evaluation, Chen2024RareBenchCL}. The input is the free-text clinical presentation, and no chain-of-thought inference is implemented;
\textbf{(ii) training-free RAG prompting}. 
By adopting prompts that equip clinical LLMs to interact with the retrieval corpus at will, the model can further integrate domain-specific knowledge during inference~\cite{xiong2024benchmarking, Xu2024BMRetrieverTL, Delile2024GraphBasedRC}. For fair comparison, this inference-only~(no traininig) setting employs the same prompt design and tool access as our agentic system but relies solely on in-context learning without any reward-based optimization;
\textbf{(iii) supervised fine-tuning~(SFT)}. 
By further fine-tune the LLM using the training dataset and cases from the patient record database in a generative way, we can align latest LLMs with clinical tasks~\cite{wu2025towards, dou2025baichuan, wu2024pmc}. This baseline establishes the performance ceiling achievable through traditional supervised learning on the same data distribution used for our reinforcement learning approach, tested on the identical evaluation set;
\textbf{(iv) target-only RL training}. 
This is a variant of our method, employing reinforcement learning but removes the specialized policy reward that guides the optimization of the reasoning and retrieval processes. Supervision is provided based solely on target outputs, 
using the same environment and parameters as our full method for ablation study on the value of process-oriented guidance.

\noindent \textbf{Other clinical diagnosis models.}
We further compare \ModelName{} against established state-of-the-art diagnostic systems:
\textbf{(i) general-purpose large language models}. 
We adopt the Qwen2.5~\cite{Yang2024Qwen25TR} and Llama3.1~\cite{dubey2024llama} series as our primary backbones. Specifically, considering the computational cost, we use \texttt{Qwen2.5-7B-Instruct}, \texttt{Qwen2.5-14B-Instruct}, and \texttt{Llama3.1-8B-Instruct}. For high-capability baselines, we employ GPT-4o~(proprietary)~\cite{hurst2024gpt} and DeepSeek-R1~(open-source)~\cite{guo2025deepseek}, accessing their official APIs with the models \texttt{DeepSeek-R1-0528} and \texttt{gpt-4o-2024-11-20};
\textbf{(ii) biomedical CLIP-based encoders}. We adopt MedCPT~\cite{Jin2023BioCPTCP}, a contrastive learning approach that treats the clinical presentation as a query to retrieve the most likely diagnosis, using their official checkpoint; \textbf{(iii) medical large language \& foundation models}. This category groups domain-adapted models trained on medical corpora~\cite{singhal2023large} and multi-modal foundation models~\cite{Liu2025AGM}. We pick Meditron~\cite{Chen2023MEDITRON70BSM} and MedGemma~\cite{sellergren2025medgemma} as representative models with strong instruction-following capabilities, using their official checkpoints; \textbf{(iv) medical RAG-based frameworks}. We include the MedRAG~\cite{xiong2024benchmarking} framework, which relies on a general medical knowledge corpus specified via system prompts without fine-tuning, using their official implementation; \textbf{(v) chain-of-thought agentic models}. We evaluate CoD~\cite{chen2025cod}, a model that incorporates chain-of-thought paradigm through supervised fine-tuning, using their official checkpoint; \textbf{(vi) multi-agent consultation systems}. We test MAC~\cite{chen2025enhancing}, a framework employing multiple role-playing agents to simulate expert consultation with their official implementation; \textbf{(vii) multi-agent systems trained with reinforcement learning}. We include DoctorAgent~\cite{feng2025doctoragent}, a system that combines multi-agent interaction with reinforcement learning and supervised fine-tuning, using their official implementation.

\subsection{Datasets}

\noindent \textbf{Medical retrieval corpus.}  
The model is augmented by a comprehensive retrieval corpus, which comprises three distinct knowledge sources: 
(1) structured disease guidelines mapping over 250k disease-phenotype pairs to standard terminologies; 
(2) a patient record database containing 155k validated clinical cases for similar-case retrieval; 
and (3) a massive clinical knowledge collection spanning millions of PubMed articles and biomedical documents from Wikipaedia to provide broad semantic context (see Sec.~\ref{sec:data_description} for more details).

\noindent \textbf{Training \& evaluation datasets.} 
To rigorously evaluate \ModelName{}, we curated a diverse multi-center cohort of over 24k clinical cases spanning seven international sources from America, Asia, and Europe. This collection balances common pathologies (73.1\%) with a significant representation of rare diseases (26.9\%), covering over 3,000 distinct diseases with high phenotypic complexity (averaging 4–12 symptoms per case). We adopted a robust validation strategy to assess both internal consistency and external generalization. Five core datasets (MIMIC-IV~\cite{Johnson2023MIMICIVAF}, PMC-Patients~\cite{Zhao2022PMCPatientsAL}, MedDialog~\cite{Chen2020MedDialogAL}, RareArena\footnote{https://huggingface.co/datasets/THUMedInfo/RareArena}, and RareBench~\cite{Chen2024RareBenchCL}) were partitioned at a 3:1 ratio to form the in-distribution training and test sets.
Crucially, to measure the model's ability to generalize to unseen clinical environments, we reserved the Mendeley~\cite{Shafi2025ASD} and Xinhua-Rare~\cite{Zhao2025AnAS} datasets exclusively for out-of-distribution zero-shot evaluation (see Extended Data Fig.~\ref{fig:datastatistics_extend} and Sec.~\ref{sec:data_description} for detailed dataset statistics and pre-processing protocols).

\subsection{Evaluation Settings}
\label{sec:eval_setting}

In this section, we define the main metrics for model evaluation and interpretability of the RAG policy.

\noindent \textbf{Performance metrics.}
We employ four quantitative metrics to assess system performance:
\textbf{(i) top‑N accuracy (Acc@N)}: measures if the ground-truth diagnosis appears within the top‑N predictions.
\textbf{(ii) Hit@N:} evaluates the retrieval policy; a ``hit'' occurs if any of the top‑N retrieved patients have the same diagnosis as ground-truth
\textbf{(iii) hint score:} the proportion of diagnostic workflows where the ground-truth disease is explicitly mentioned during reasoning, even if not the final prediction;
\textbf{(iv) retrieval action steps:} The count of external information seeking actions ($\langle\text{\texttt{lookup}}\rangle$, $\langle\text{\texttt{match}}\rangle$, $\langle\text{\texttt{search}}\rangle$) per trajectory, quantifying the extent of evidence acquisition.

\noindent \textbf{Assessment on reasoning~(LLM-as-a-Judge).}
To scale the evaluation of reasoning quality, we implemented an automated ``LLM-as-a-judge'' protocol~\cite{croxford2025evaluating}. We adopted two foundation models: DeepSeek-R1~\cite{guo2025deepseek} (for general reasoning) and Meditron-70B~\cite{chen2023meditron} (for domain knowledge). 
These models scored cases across the five holistic dimensions defined in Tab.~\ref{tab:holistic-wise}, as defined by collaborating with expert clinician~(8-year experience in both clinical practice and AI).  To ensure robustness, scores were averaged across three independent runs with these two LLMs. The system prompt used to instruct these evaluators is provided in supplementary Sec.~\ref{supp:reasoning_assess_prompt}.

\begin{table}[t]
\centering
\scriptsize
\caption{Holistic evaluation framework for clinical reasoning (Score 1--4). }
\label{tab:holistic-wise}
\begin{tabularx}{\textwidth}{@{}l l X@{}}
\toprule
\textbf{Dimension} & \textbf{Score} & \textbf{Criteria Description} \\
\midrule
\textbf{Overall Correctness} & \textit{1--4} & From incorrect/dangerous (1) to precise identification of ground truth and exclusions (4). \\
\midrule
\textbf{Clinical Utility} & \textit{1--4} & From disorganized/no insight (1) to consultant-level analysis and management (4). \\
\midrule
\textbf{Reasoning Consistency} & \textit{1--4} & From chaotic/fallacious logic (1) to complex synthesis and robust coherence (4). \\
\midrule
\textbf{Diagnostic Filedity} & \textit{1--4} & From off-topic/irrelevant (1) to highly pertinent, high signal-to-noise ratio (4). \\
\midrule
\textbf{Hallucination Severity} & \textit{1--4} & From severe fabrication of symptoms (1) to completely faithful to case description (4). \\
\bottomrule
\end{tabularx}
\end{table}

\noindent \textbf{Interpretability of RAG policy.}
We analyzed the learned RAG policy using four indicators:
\textbf{(i) symptom association:} proxied by Hit@20, this measures the model's ability to extract pertinent clinical features that lead to relevant case retrieval;
\textbf{(ii) differential diagnosis:} measured by top-5 accuracy when the model is provided with retrieved context containing both the ground-truth and differentials, assessing discriminative capacity;
\textbf{(iii) irrelevance exclusion:} measured by the maintenance of top-5 accuracy when the retrieval module may return irrelevant noise, quantifying robustness against distraction;
\textbf{(iv) reasoning complexity:} quantified by the average number of action steps per trajectory, serving as a proxy for the thoroughness of the diagnostic investigation.

\begin{figure}[!t]
    \centering
    \includegraphics[width=1\linewidth]{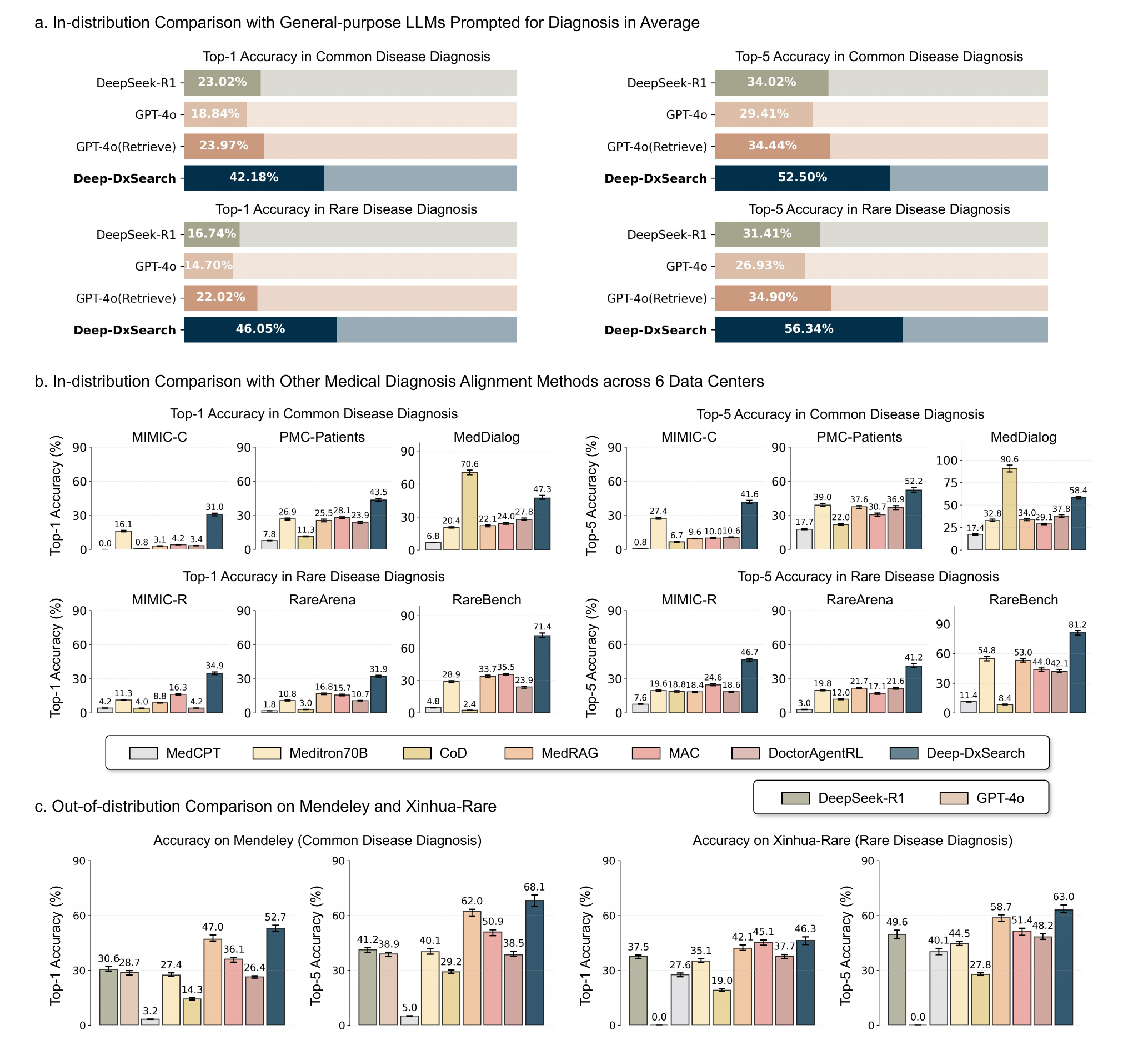}
    \caption{\textbf{Comparison with previous diagnostic methods.} \textbf{a,} Comparison of \ModelName{} with general-purpose LLMs---including GPT-4o, GPT-4o with retrieval, and DeepSeek-R1---on common and rare disease diagnosis (averaged across in-distribution datasets). \textbf{b,} Detailed performance breakdown of \ModelName{} versus medical-specific systems across individual in-distribution datasets. \textbf{c,} Comparative evaluation of \ModelName{} against all these diagnostic methods on out-of-distribution (OOD) datasets. Note, GPT-4o was excluded on Xinhua-Rare due to privacy constraints associated with this in-house dataset.}
    \label{fig:DiagCompare}
\end{figure}

\subsection{Deep-DxSearch Outperforms State-of-the-Art Models in Multi-Center Benchmarks}
\label{sec:sota_comparison}

We evaluated diagnostic accuracy across the multi-center cohort, implementing \ModelName{} with the \textbf{MedGemma-27B} backbone. The system demonstrated consistent superiority over both general-purpose reasoners and specialized medical frameworks.

\noindent \textbf{In-distribution (ID) evaluation.}
On common disease datasets, \ModelName{} achieved an average Top-1 accuracy of 42.2\%, substantially outperforming DeepSeek-R1 (23.0\%) and GPT-4o (18.8\%) (Fig.~\ref{fig:DiagCompare}a). Notably, simply augmenting GPT-4o with our retrieval corpus (RAG-only) yielded only 24.0\%, highlighting that access to data alone is insufficient without a learned retrieval policy.
For rare diseases—a critical stress test for diagnostic AI—\ModelName{} attained 52.5\% top-1 accuracy, widening the gap against DeepSeek-R1 (34.0\%) and GPT-4o (29.4\%). Among medical-specific systems (Fig.~\ref{fig:DiagCompare}c), our framework consistently led the field, surpassing the strong Meditron-70B baseline by over 20 percentage points in common disease accuracy and outperforming the multi-agent MAC system in rare disease scenarios.

\noindent \textbf{Out-of-distribution (OOD) evaluation.}
The robustness of the learned policy was evidenced by zero-shot performance on unseen datasets~(Fig.~\ref{fig:DiagCompare}d). On the Mendeley benchmark, \ModelName{} achieved 52.7\% top-1 accuracy, surpassing the second-best method (MedRAG) by 5.8\%. Similarly, on the Xinhua-Rare dataset, it attained 46.3\% accuracy, outperforming MAC (45.1\%) and MedRAG (35.8\%). These results indicate that the agentic reasoning patterns learned via RL transfer effectively to novel clinical environments, maintaining top-5 accuracy consistently above 60\%.

\subsection{Process-Based Rewards and Structured Retrieval Drive Performance Gains}
\label{sec:ablation}

To deconstruct the performance gains, we analyzed the contributions of the end-to-end RL training paradigm, the reward structure, and the retrieval corpus.

\begin{figure}[!htb]
    \centering
    \includegraphics[width=1\linewidth]{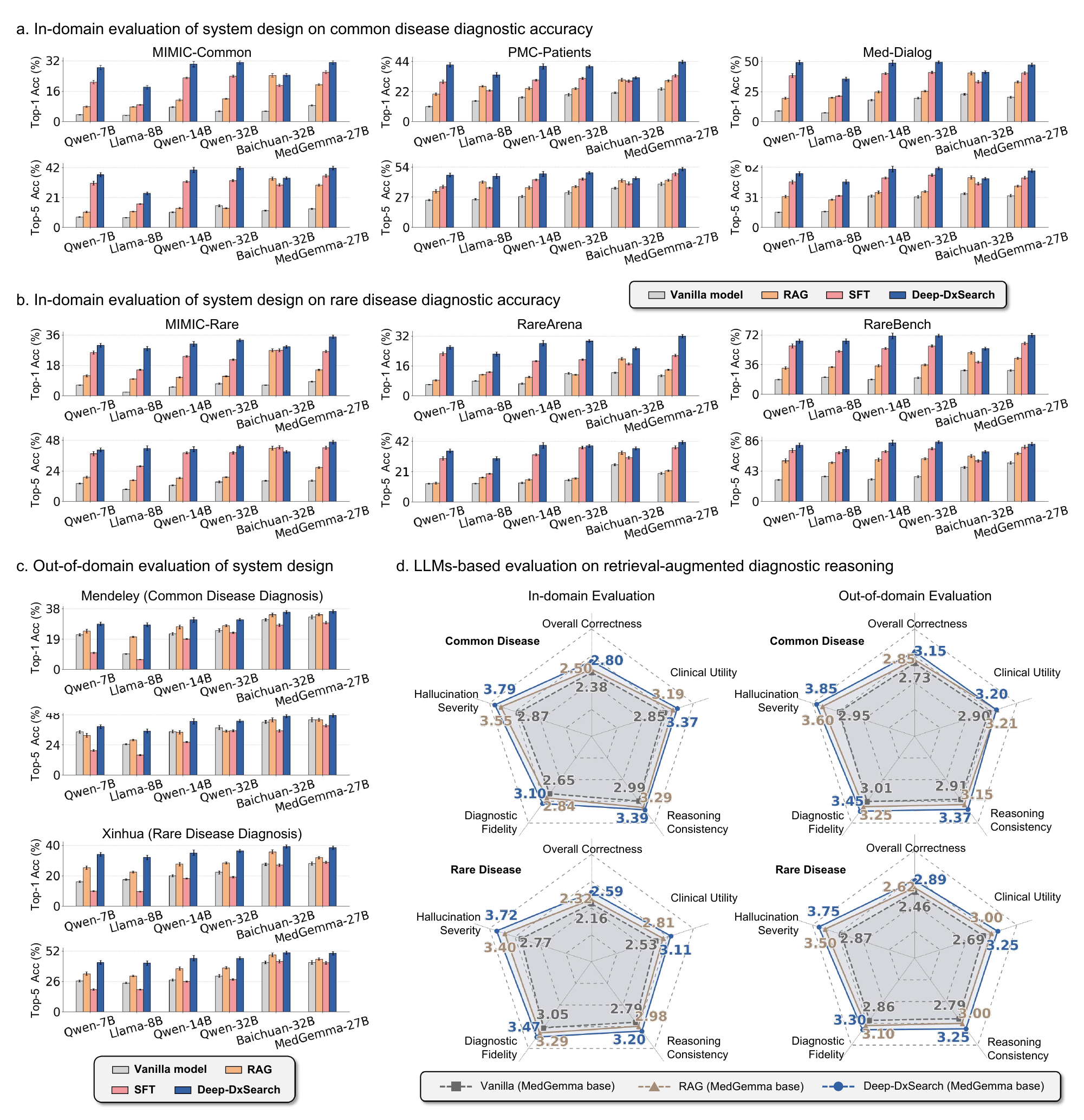}
    \caption{\textbf{Evaluation of \ModelName{} system design.} \textbf{a}. In-distribution evaluation on common disease diagnosis comparing top-1 and top-5 accuracy among vanilla, RAG, SFT, and \ModelName{} approaches. 
    \textbf{b}. In-distribution evaluation on rare disease diagnosis following identical settings. 
    \textbf{c}. Out-of-distribution evaluation covering both common and rare diseases. 
    \textbf{d}. Assessment of diagnostic reasoning quality using an ``LLM-as-a-judge'' protocol across five clinical dimensions. Scores range from 1 to 4; notably, hallucination severity is scored inversely, such that a score of 4 indicates the minimal degree of hallucination.}
    \label{fig:SystemDesignComparison}
\end{figure}

\noindent \textbf{Impact of end-to-end policy optimization with reinforcement learning.}
We compared the full \ModelName{} framework against vanilla LLM, training-free RAG, and Supervised Fine-Tuning (SFT). Across all tested backbones~(Qwen, Llama, Baichuan, MedGemma), the RL-trained agent consistently yielded the highest accuracy~(Fig.~\ref{fig:SystemDesignComparison}a-c). For instance, on the MedDialog dataset, Qwen7B trained with \ModelName{} achieves 49.3\% accuracy, surpassing vanilla (9.0\%) and RAG (19.6\%); similarly, on the Xinhua-Rare dataset, BaichuanM2 trained via \ModelName{} attains a top-1 accuracy of 39.3\%, exceeding the vanilla (27.6\%) and RAG (35.8\%), respectively. Crucially, \ModelName{} addresses the overfitting observed in supervised fine-tuning~(SFT), while SFT improves in-distribution performance ({\em e.g.}, MedGemma on MIMIC-R improves from 16.4\% to 42.3\%), it often degrades on out-of-distribution benchmarks due to overfitting~({\em e.g.}, Llama8B drops from 17.5\% to 9.6\% on Xinhua-Rare). In contrast, \ModelName{} maintains the strong performance on out-of-distribution benchmarks. 

Additionally, using an ``LLM-as-a-judge'' protocol (DeepSeek-R1 and Meditron-70B)~(detailed in Sec.~\ref{sec:eval_setting}), we find that \ModelName{} achieves superior reasoning quality compared to vanilla and RAG baselines~(Fig.~\ref{fig:SystemDesignComparison}d). 
Specifically, on OOD common disease settings, \ModelName{} attains a \textit{reasoning consistency} score of 3.37, 
surpassing both vanilla (2.91) and RAG (3.15) methods. 
For OOD rare diseases, the model demonstrates enhanced safety, achieving a \textit{hallucination severity} score of 3.75, outperforming the vanilla baseline (2.87) and improving upon standard RAG (3.50).

\begin{figure}[!h]
    \centering
    \includegraphics[width=1\linewidth]{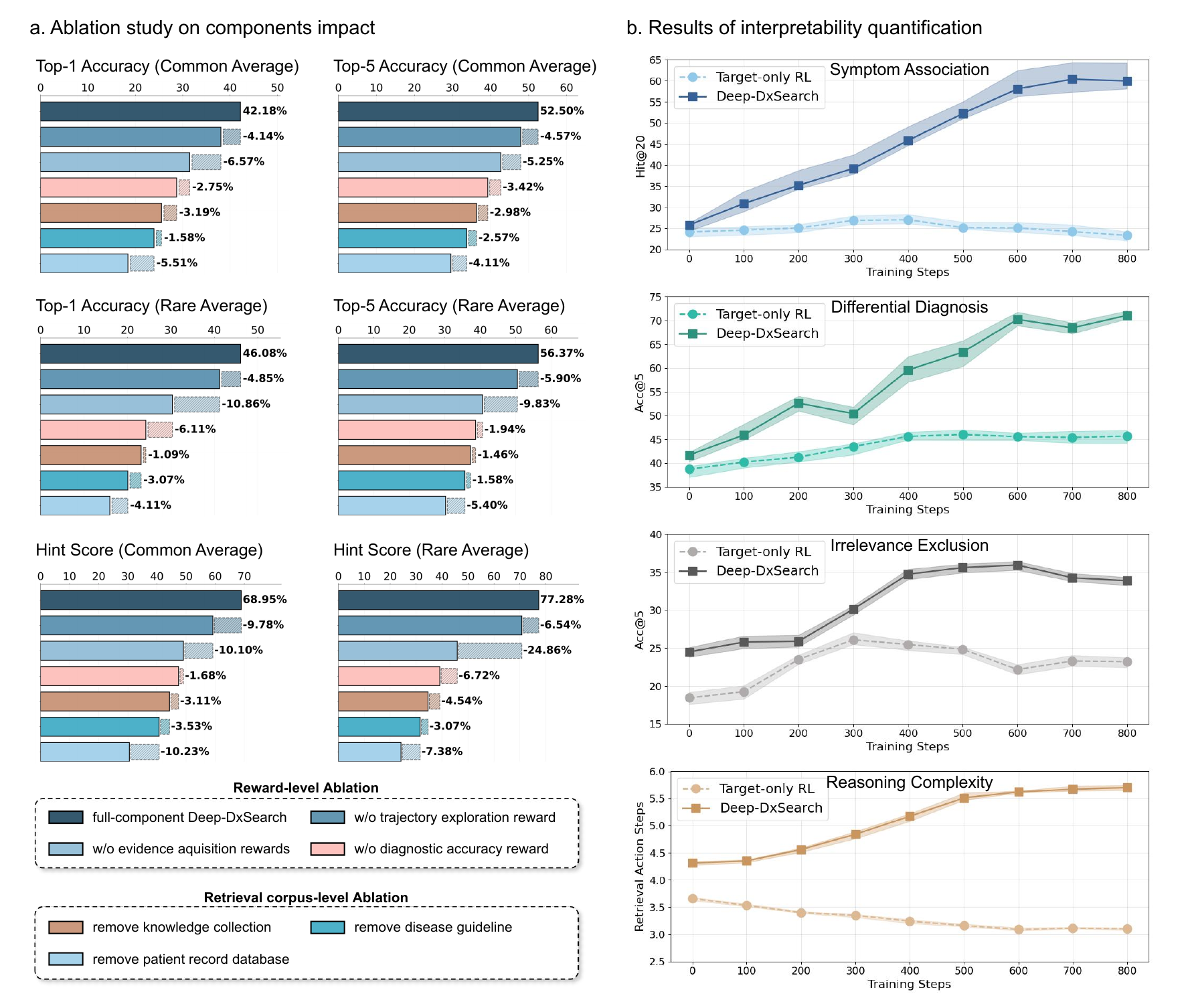}
    \caption{\textbf{Ablation study and interpretability analysis.} 
    \textbf{a}, Performance variation following stepwise removal of \ModelName{}'s components. The top section evaluates reward designs: trajectory exploration, retrieve-reason, and diagnostic target rewards. The bottom section evaluates the retrieval corpus: document summarizer, knowledge collection, disease guidelines, and patient records. ``Hint'' denotes cases where the correct disease is considered during reasoning. 
    \textbf{b}, Quantitative metrics for diagnostic policy. Panels display: ``Symptom association'' (hit@20 for case retrieval); ``Differential diagnosis'' (top-5 accuracy); ``Irrelevance exclusion'' (robustness to misleading evidence); and ``Reasoning complexity'' (average action steps). \ModelName{} is compared against a ``target-only'' baseline trained without intermediate supervision.}
    \label{fig:AblationInterpret}
\end{figure}

\noindent \textbf{Ablation on reinforcement learning rewards.}
We validate our composite reward function by progressively removing components~(Fig.~\ref{fig:AblationInterpret}a):
\textbf{(i) removing the trajectory exploration reward} reduces accuracy by $\sim$4\% across datasets, resulting in rigid diagnostic workflows;
\textbf{(ii) removing the evidence acquisition rewards} causes a sharp decline on accuracy~(6.6\% common, 10.9\% rare), confirming the need to incentivize valid intermediate steps;
\textbf{(iii) removing the diagnostic accuracy reward} causes a further drop on accuracy~($\sim$2--6\%). This ablation suggests that while the final diagnostic target provides the primary directional signal, the retrieve-reason incentive is the most critical driver for navigating the complex search space, with trajectory exploration serving as a necessary regularizer to prevent policy collapse.

\noindent \textbf{Ablation on the retrieval corpus.}
We evaluate the impact of the retrieval corpus by progressively removing components from the environment~(Fig.~\ref{fig:AblationInterpret}a):
\textbf{(i) removing the clinical knowledge collection} results in a net accuracy decline (3.2\% common, 1.1\% rare);
\textbf{(ii) removing structured guidelines} leads to consistent decreases (1.6\% common, 3.1\% rare), validating their supportive function;
\textbf{(iii) excluding patient record matching} yields the most substantial performance drop (5.5\% common, 4.1\% rare), underscoring the dominant role of evidence-based matching~(see Extended Data Tab.~\ref{tab:embedding_ablation_extend} for more details about the granular progressive removal analysis).
Overall, these results indicate that while similar-case retrieval serves as the cornerstone of diagnostic accuracy, the system relies on the synergistic integration of summarized knowledge and structured guidelines to maximize precision and coverage. For further analysis regarding the impact of these retrieval actions on temporal efficiency, please refer to Extended Data Tab.~\ref{tab:action_ablation_extend}.

\noindent \textbf{Evolution of diagnostic policy.}
As shown in Fig.~\ref{fig:AblationInterpret}b, 
we also tracked the evolution of \ModelName{}'s policy~(as explained in Sec.~\ref{sec:eval_setting}) when comparing to a target-only baseline, \ModelName{} demonstrates: 
\textbf{(i) improved symptom association:} 
\textit{hit@20}—defined as the proportion of cases where the top-20 retrieved patient records contain the ground-truth diagnosis—rises from 25.8\% to 60.4\%; 
\textbf{(ii) differential diagnosis:} we assess the ability to identify the correct diagnosis among candidates using top-5 accuracy. 
While the baseline improves moderately from 38.7\% to 45\%, \ModelName{} achieves significant increase to 71.1\%, 
indicating a stronger capability to distinguish the correct pathology from plausible alternatives;
\textbf{(iii) robustness:} while injecting with irrelevant documents during test time, \ModelName{} maintains a 10\% accuracy gain over the baseline; 
\textbf{(iv) reasoning complexity:} the average trajectory length increases to $>$5.5 steps, whereas the baseline shrinks to $\sim$3 steps, indicating that \ModelName{} learns to engage in comprehensive iterative reasoning rather than guessing. Extended Data Fig.~\ref{fig:transition_extend} presents further results regarding the evolution of \ModelName{}'s diagnostic policy.

\subsection{Physicians Prefer Deep-DxSearch for Transparent Decision Support}
\label{sec:human_eval}

To assess clinical utility, we conducted a \textbf{physician-in-the-loop study} involving three clinicians (junior, medium, senior). The evaluation focused on two dimensions: the impact on diagnostic accuracy and the transparency of the reasoning process.

\begin{figure}[t]
    \centering
    \includegraphics[width=1\linewidth]{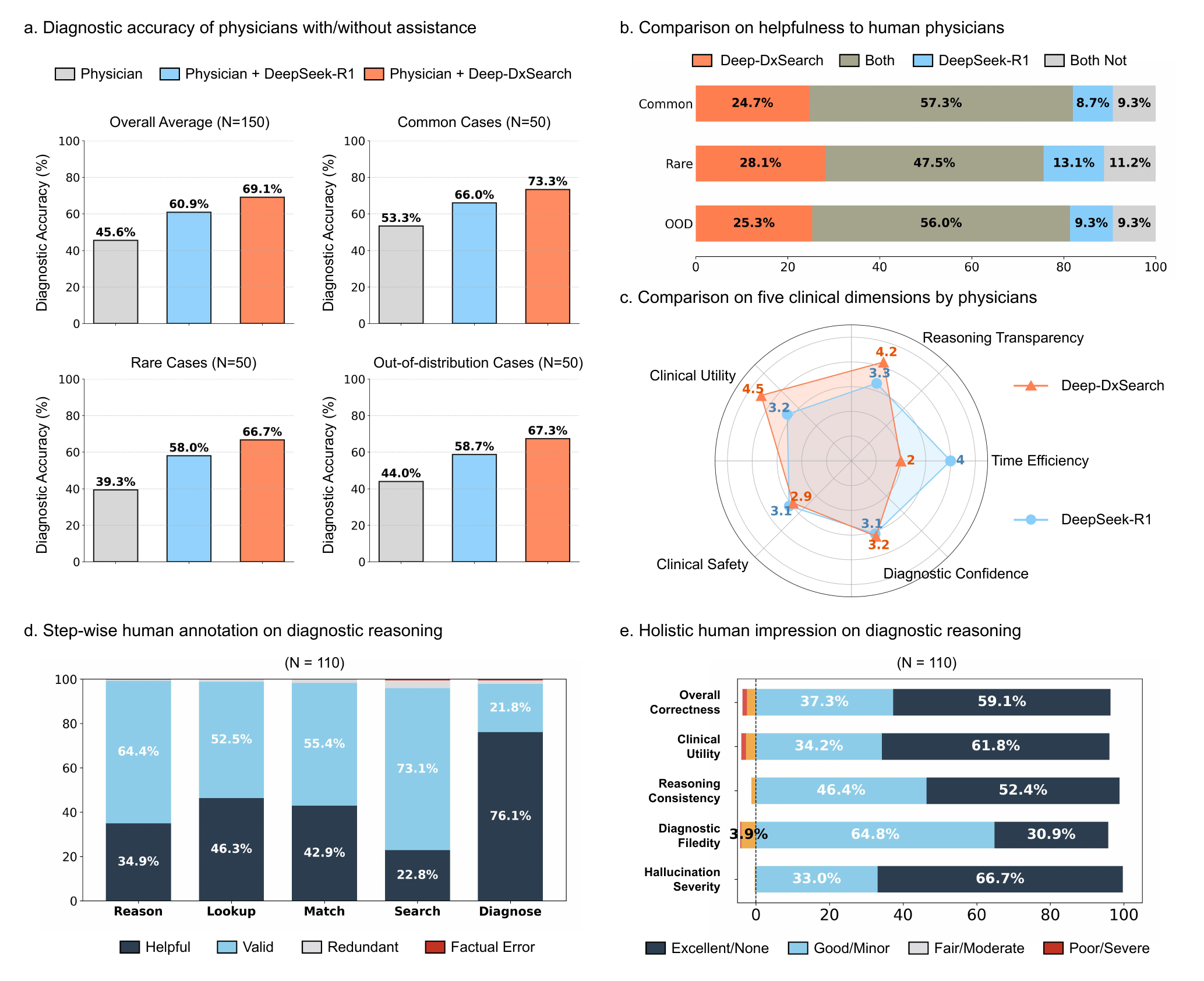}
    \caption{\textbf{Physician-involved Evaluation Results.} 
    \textbf{a}, Diagnostic accuracy comparison under unaided, DeepSeek-R1-aided, and \ModelName{}-aided conditions ($N=150$). 
    \textbf{b}, Physician preference distribution across dataset types. 
    \textbf{c}, Multidimensional clinical assessment of AI assistants. 
    \textbf{d}, Expert adjudication of step-wise action validity ($N=110$). 
    \textbf{e}, Holistic evaluation of full diagnostic trajectories across correctness, utility, and hallucination.}
    \label{fig:HumanEval}
\end{figure}

\noindent \textbf{Case preparations.}
To facilitate the annotation process, 
we developed a custom web interface (based on Gradio) that presents extensive clinical data (EHRs) alongside \ModelName{}'s reasoning trajectories. The interface delineates critical reasoning steps for assessment and allows evaluators to navigate and annotate the diagnostic workflow flexibly, ensuring a streamlined review process.

We collaborated with an experienced clinician from Shanghai Ninth People's Hospital, Shanghai Jiao Tong University School of Medicine, with 8 years of clinical experience and an AI background, 
to curate a total of 260 test cases~(with both common and rare disease), drawn from both in-distribution and out-of-distribution sources. All diagnostic labels for these records were strictly excluded and manually verified to prevent information leakage. 
This collection was split into two distinct subsets for evaluation. 
The first subset ($N=150$), designed for the \textbf{AI assistance study}, consists of raw EHRs to represent authentic clinical encounters. It comprises 50 common and 50 rare cases from in-distribution datasets, along with 50 out-of-distribution cases from the Xinhua-Rare dataset. 
The second subset ($N=110$), spanning both common and rare diseases, 
was employed for the \textbf{reasoning evaluation}. 
For these cases, we preserved the complete inference trajectories, including inputs, intermediate retrieval-reasoning steps, and ground truth, to facilitate granular expert adjudication.

\noindent \textbf{Study-I: impact on AI assistance in clinical decision-making.}
Three physicians—senior~($>$10 years), medium~(8 years), and junior~(6 years)—performed diagnoses under \textbf{unassisted} and \textbf{AI-assisted} conditions (using DeepSeek-R1 vs.~\ModelName{}). 
To minimize carryover effects, the assistance order was randomized, and physicians were required to justify rational of their model preferences.
As shown in Fig.~\ref{fig:HumanEval}a, the average physician accuracy reached 69.1\% when aided by {\ModelName{}}, exceeding the 45.6\% unassisted baseline and the 60.9\% achieved with DeepSeek-R1. Notably, for OOD cases, \ModelName{} assistance yielded an accuracy of 67.3\% compared to 44.0\% (unassisted) and 58.7\% (DeepSeek-R1).
Physicians also rated their preference on common, rare, and OOD datasets~(Fig.~\ref{fig:HumanEval}b), where {\ModelName{}} clearly shows stronger preference than DeepSeek-R1 or without AI assistance. 
In a multi-dimensional assessment with metrics shown in Tab.~\ref{tab:human-eval-rubric}, \ModelName{} scored significantly higher than DeepSeek-R1 in \textbf{reasoning transparency} (4.2 vs. 3.3) and \textbf{clinical utility} (4.5 vs. 3.2), although it received lower scores for \textbf{time efficiency} (2.0 vs. 4.0) due to the time load required to review the retrieved evidence evidence, as shown in Fig.~\ref{fig:HumanEval}c.

\begin{table}[t]
\centering
\scriptsize
\caption{Five-dimensional evaluation framework for AI-assisted clinical diagnosis (Score 1--5) with ``3'' denotes the medium level. }
\label{tab:human-eval-rubric}
\begin{tabularx}{\textwidth}{@{}l l X@{}}
\toprule
\textbf{Dimension} & \textbf{Score} & \textbf{Criteria Description} \\
\midrule
\multirow{2}{*}{\textbf{Reasoning Transparency}} 
 & \textit{Low (1-2)} & Opaque logic; contradictions; vague or hallucinated evidence. \\
 & \textit{High (4-5)} & Clear reasoning steps; perfect evidence chain anchored to verifiable sources. \\
\midrule
\multirow{2}{*}{\textbf{Clinical Utility}} 
 & \textit{Low (1-2)} & Irrelevant noise; generic textbook knowledge without patient specificity. \\
 & \textit{High (4-5)} & Highlights overlooked symptoms; provides consultant-level insight that optimizes decisions. \\
\midrule
\multirow{2}{*}{\textbf{Diagnostic Confidence}} 
 & \textit{Low (1-2)} & Physician rejects advice or requires rigorous cross-verification. \\
 & \textit{High (4-5)} & Physician trusts the output; willing to adopt diagnosis as primary basis. \\
\midrule
\multirow{2}{*}{\textbf{Clinical Safety}} 
 & \textit{Low (1-2)} & Missed life-threatening conditions; contraindicated actions; major errors. \\
 & \textit{High (4-5)} & Safe; compliant with standard care; actively flags contraindications. \\
\midrule
\multirow{2}{*}{\textbf{Time Efficiency}} 
 & \textit{Low (1-2)} & System response $>$ 30 seconds. \\
 & \textit{High (4-5)} & System response $<$ 20 seconds. \\
\bottomrule
\end{tabularx}
\end{table}


\noindent \textbf{Study-II: evaluation of diagnostic reasoning steps.} 
To validate the intermediate reasoning process, the model's outputs 
of the sampled 110 cases are decomposed into 942 distinct action steps (\texttt{<reason>}, \texttt{<lookup>}, \texttt{<match>}, \texttt{<search>}, \texttt{<diagnose>}). Three independent clinicians~(junior, medium, senior) evaluated these steps as either ``factual error'', ``redundant''~(correct but unnecessary), ``valid''~(logical and grounded), or ``helpful''~(key reference). 
Results~(Fig.~\ref{fig:HumanEval}d) indicate a high density of useful information: steps rated as ``helpful'' comprised 34.9\% of \texttt{<reason>} steps, 46.3\% of \texttt{<lookup>}, and 76.1\% of \texttt{<diagnose>} steps. The combined proportion of ``valid'' and ``helpful'' steps reached 99.3\% for \texttt{<reason>} 
and 98.8\% for \texttt{<lookup>}.

\noindent \textbf{Study-III: holistic reasoning evaluation.}
Physicians further evaluated full diagnostic trajectories with the same rubric as ``LLM-as-judge''. As shown in Fig.~\ref{fig:HumanEval}e, \ModelName{} was rated as ``excellent'' in overall correctness for 59.1\% of cases. Additionally, 61.8\% of trajectories were deemed to have high clinical utility, and 66.7\% were considered {\em free from} hallucinations. To strictly validate the reliability of these evaluations, we calculated the \textbf{positive percent agreement~(PPA)} averaged among three physicians based on binarized outcomes (positive: score 3--4 vs.~negative: score 1--2). 
The analysis reveals a strong consensus among evaluators, yielding PPA values of 96.7\% for overall correctness, 96.1\% for clinical utility, 98.7\% for reasoning consistency, 95.5\% for reference information relevance, and 99.7\% for hallucination severity.

\begin{figure}[t]
    \centering
    \includegraphics[width=1\linewidth]{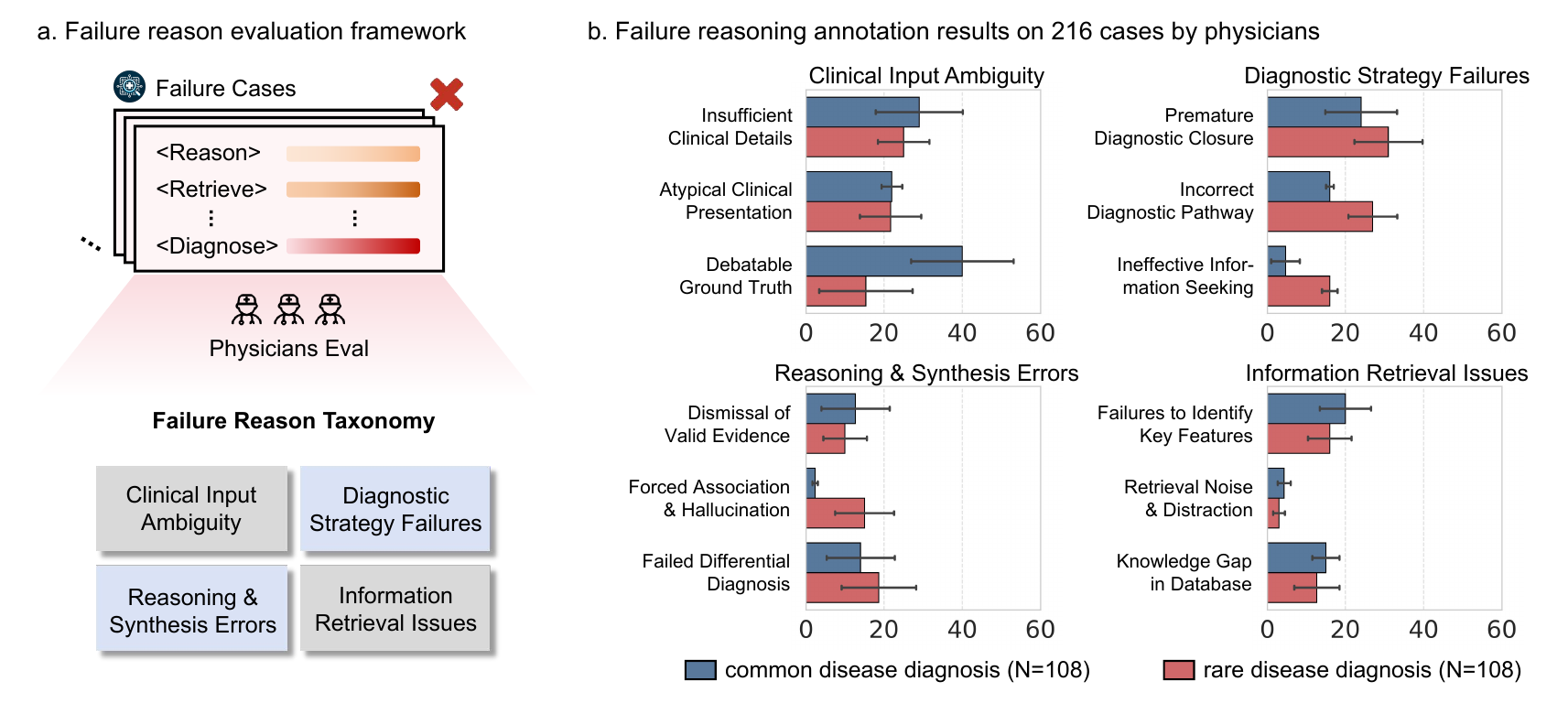}
    \caption{\textbf{Failure Analysis.} \textbf{a}, Schematic of the physician-led evaluation framework and taxonomy used to categorize diagnostic errors into four primary domains. \textbf{b}, Quantitative comparison of failure reasons for common ($N=108$) and rare ($N=108$) disease cases across twelve specific error types. Error bars represent the deviation among the three physician evaluators.}
    \label{fig:failurecase}
\end{figure}

\subsection{Failure Analysis}

To elucidate the limitations, we conducted a failure analysis on 216 cases—defined as instances where the correct diagnosis was absent from the model's top-5 predictions—stratified evenly between common ($n=108$) and rare ($n=108$) diseases. We established a taxonomy of failure modes comprising four domains subdivided into 12 specific categories: \textbf{(i) clinical input ambiguity}~(including insufficient details, atypical presentation, and debatable ground truth); 
\textbf{(ii) information retrieval issues}~(failure to identify key features, retrieval noise, and database knowledge gaps); 
\textbf{(iii) diagnostic strategy failures}~(premature closure, incorrect pathway, and ineffective information seeking); 
and \textbf{(iv) reasoning \& synthesis errors}~(dismissal of valid evidence, hallucination, and failed differential diagnosis). Three external physicians with varying seniority (6 to $>$10 years of experience across three tertiary hospitals) adjudicated the primary factor(s) contributing to the diagnostic failure for each case (Fig.~\ref{fig:failurecase}a).

\noindent \textbf{Failure pattern in common disease diagnosis.}
Failures in common diseases stemmed primarily from clinical input ambiguity and retrieval issues~(Fig.~\ref{fig:failurecase}b). Notably, ``debatable ground truth'' accounted for 40 cases (vs.~15.3 in rare diseases), suggesting the model's diagnosis was often clinically justifiable despite differing from the label. Similarly, ``failure to identify key features'' was the dominant retrieval error (20 vs.~15 cases), indicating challenges in prioritizing symptoms.

\noindent \textbf{Failure pattern in rare disease diagnosis.}
Conversely, failures in rare disease were driven by diagnostic strategy and reasoning errors. ``Premature diagnostic closure''~(finalizing diagnosis without sufficient evidence) occurred in 31 cases, surpassing the 24 observed in common diseases. Additionally, ``failed differential diagnosis''~(errors in distinguishing competing hypotheses) was identified in 18.7 cases (vs.~12 for common), highlighting deficits in managing complex knowledge.

These results reveal distinct failure modes: common disease diagnosis is hindered by input noise and retrieval precision, whereas rare disease diagnosis suffers from strategic reasoning deficits. This divergence explains the observed performance disparities and underscores the need for both cleaner retrieval corpora and robust reasoning synthesis.

\clearpage
\section{DISCUSSION}

The integration of artificial intelligence into clinical diagnostics has historically been constrained by a dichotomy: the ``black box'' opacity of deep learning versus the hallucination risks inherent in generative large language models (LLMs). In this study, we introduce \ModelName{}, a framework that transitions diagnostic AI from static classification to dynamic, agentic inquiry. By coupling a massive medical retrieval environment with a policy optimized via reinforcement learning, our system demonstrates that mimicking the hypothetico-deductive process of human clinicians—rather than merely predicting the next token—yields superior diagnostic accuracy and generalization.

\noindent \textbf{From pattern matching to deliberative reasoning}.
A critical finding of our work is the limitation of standard Supervised Fine-Tuning (SFT) for complex diagnostics. While SFT improves in-distribution performance, our results show it often degrades on out-of-distribution cohorts ({\em e.g.}, Xinhua-Rare), suggesting a tendency to memorize specific disease-symptom associations. In contrast, \ModelName{}, trained via agentic reinforcement learning, learns a transferable \textit{diagnostic policy}. This aligns with the cognitive distinction between System 1 (fast, intuitive) and System 2 (slow, deliberative) thinking. By explicitly rewarding the acquisition of valid evidence and the exploration of diverse search trajectories, \ModelName{} effectively implements evidence-based reasoning. This allows the agent to navigate unseen clinical environments with a robustness that standard RAG and SFT approaches lack, bridging the gap between rigid heuristic workflows and purely generative models.

\noindent \textbf{Addressing the long-tail of rare diseases}.
Rare diseases represent a distinct challenge due to their low prevalence and phenotypic complexity, often resulting in a ``diagnostic odyssey'' for patients. Existing benchmarks, such as the RareArena dataset, highlight the struggle of current foundation models to identify these conditions due to the sparsity of training data. \ModelName{} significantly advances this frontier, achieving substantial accuracy gains over strong baselines like GPT-4o and DeepSeek-R1 in rare disease cohorts. This performance leap is attributable to the system's ability to anchor reasoning in retrieved, validated cases via the \texttt{<match>} primitive. By referencing a database of over 170k patient records rather than relying solely on parametric memory, our model mitigates the ``long-tail'' forgetting problem. However, our failure analysis indicates that challenges remain; specifically, the model occasionally exhibits ``premature closure'', finalizing diagnoses before fully exploring latent pathologies. This suggests that future iterations must enforce even more rigorous differential diagnosis protocols before the agent is permitted to conclude a case.

\noindent \textbf{Clinical utility and the transparency trade-off}.
The ultimate measure of a diagnostic tool is its utility in human-AI collaboration. Our physician-in-the-loop study demonstrates that \ModelName{} not only improves diagnostic accuracy (69.1\% assisted vs.~45.6\% unassisted) but also fosters trust through transparency. Unlike opaque predictions, the explicit action steps (e.g., \texttt{<lookup>}, \texttt{<search>}) provide an audit trail that physicians rated highly for reasoning transparency. It is notable that this deliberative process resulted in lower ratings for time efficiency compared to direct-answer models. In the context of high-stakes medicine, we argue this is a necessary trade-off: the ``glass box'' nature of \ModelName{} offers a safeguard, allowing clinicians to verify the evidence chain before accepting an AI's conclusion. This shifts the role of the AI from an oracle to a transparent research assistant, aligning with the requirements of evidence-based medicine.

\noindent \textbf{Limitations and future directions}.
Our study has limitations that outline the path for future research. 
\textbf{First}, while the retrieval corpus is extensive, it is not exhaustive; ``database knowledge gaps'' were identified as a contributing factor to errors. Real-world viability will require mechanisms for continuous, automated updating of the knowledge base to reflect the latest biomedical literature. 
\textbf{Second}, the current modality is text-only. Clinical diagnosis is inherently multimodal, often relying on radiology, pathology, and genomics. Extending the agentic framework to process multimodal evidence is a critical next step. 
\textbf{Finally}, while hallucination severity was significantly reduced, it was not eliminated. The persistence of synthesis errors suggests that even with perfect retrieval, the reasoning engine requires further refinement to ensure logical consistency in complex scenarios.

\noindent \textbf{Conclusion}. \ModelName{} establishes a new standard for automated diagnosis by formalizing the clinical workflow as a reinforcement learning problem. By prioritizing the \textit{process} of reasoning over the immediate \textit{output}, we provide a blueprint for the next generation of medical AI: systems that are not only accurate but also transparent, verifiable, and aligned with the rigorous standards of clinical practice.

\clearpage
\section{METHODS}

In this section, we aim to provide the details for developing \ModelName{},
starting from the construction of the medical retrieval ecosystem and the automated pipeline used to curate high-fidelity clinical data. Next, we formulate the diagnostic task as a sequential decision-making process, defining the agent-environment interaction dynamics. Finally, we outline the composite reward mechanism and the reinforcement learning strategies employed to optimize the system for active, evidence-driven inquiry.

\subsection{Dataset Description and Statistics}
\label{sec:data_description}

To construct a comprehensive ecosystem comprising training datasets, evaluation benchmarks, and retrieval corpora, the curation process involved aggregating diverse clinical resources from eight primary sources, applying a rigorous automated processing pipeline, and validating data quality through expert review.

\noindent \textbf{Data Sources}

The foundation of our study rests on a diverse collection of large-scale clinical narratives used for model training and patient record retrieval. We aggregated Electronic Health Records (EHRs) and case reports from four primary sources: MIMIC-IV~\cite{Johnson2023MIMICIVAF}, comprising 332k de-identified discharge summaries from Beth Israel Deaconess Medical Center; PMC-Patients~\cite{Zhao2022PMCPatientsAL}, which provides 250k patient profiles derived from biomedical literature in PubMed Central; and MedDialog~\cite{Chen2020MedDialogAL}, containing 257k clinical consultations from online platform. To enhance rare disease coverage, we incorporated the dataset from Xinhua Hospital~\cite{Zhao2025AnAS}, a proprietary in-house collection of 352k diagnostic records spanning a decade (2014–2025). 

To support specific diagnostic tasks, we integrated specialized benchmark datasets. RareArena\footnote{\url{https://github.com/zhao-zy15/RareArena}} provided approximately 50,000 cases specifically curated for rare disease screening and confirmation. 
RareBench~\cite{Chen2024RareBenchCL} contributed 1,122 cases aggregated from RAMEDIS, MME, HMS, and LIRICAL. We also employed the Mendeley dataset~\cite{Shafi2025ASD}, a structured resource of binary disease-symptom associations released in June 2025.

Finally, to curate the retrieval corpus, we integrated structured taxonomies (ICD-10-CM, Orphanet). For unstructured evidence, we aggregated 23.9 million PubMed abstracts, 3.31 million Wikipedia entries, 18 authoritative medical textbooks, and 1,419 distinct web resources (e.g., specific sections from NCBI, NHS) for further processing. More details and licensing information are provided in Supplementary.

\noindent \textbf{Data Processing and Curation Pipeline}

Raw clinical data from the sources listed above (excluding the pre-structured Mendeley and RareBench datasets) went through a four-stage processing pipeline to ensure high fidelity and standardization.

\textbf{Stage-I: disease classification.}
Cases were split into ``common'' and ``rare'' categories based on the Orphanet ontology. A condition was classified as rare if it was listed in the Orphanet database or possessed a BioLORD~\cite{piran2024disentanglement} semantic embedding similarity of $>0.95$ with a known entry. For structured data~({\em e.g.}, MIMIC-IV), we utilized ICD-to-Orphanet cross-referencing. For unstructured datasets~({\em e.g.}, PMC-Patients, MedDialog), we implemented a rigorous annotation pipeline: candidate diagnoses were first extracted via a consensus of Large Language Models (GPT-4o and DeepSeek) and subsequently verified against the Orphanet definition via BioLORD embeddings. 
Existing rare disease benchmarks (RareArena, RareBench) retained their native classifications.

\textbf{Stage-II: pre-filtering.} 
To mitigate noise inherent in large-scale datasets, we employed GPT-4o to screen raw cases with strict criteria. We removed the ones that lack causal logic, where the diagnosis contradicted or preceded the clinical presentation; narrative incoherence, such as garbled text or administrative metadata; and information insufficiency, where the record lacked substantial clinical history. For the RareArena, RareBench and Mendeley dataset, which were pre-curated using similar protocols, we bypassed this step to avoid redundancy.

\textbf{Stage-III: extraction and stratification.} 
We adopted GPT-4o to extract confirmed diagnoses and phenotypes using a robust two-step prompting strategy to ensure fidelity and prevent information leakage. First, we employed \textit{extraction with reflection}, using Chain-of-Thought to distinguish active symptoms from medical history while pinpointing the definitive diagnosis. Second, we enforced \textit{consistency verification}, cross-referencing extracted entities against the source text to flag unsupported elements. We then applied strict quality criteria: 
(i) diseases must be well-defined clinical entities rather than vague qualifiers or simple symptoms; (ii) phenotypes must be distinctive and not merely restate the diagnosis to strictly exclude label leakage; and (iii) demographic context must align with the case. Cases satisfying these rigorous standards were designated for \textbf{training and evaluation}, while valid clinical records lacking this specific structural precision were reallocated to the patient record database to preserve real-world diversity.

\textbf{Stage-IV: terminology normalization.} 
The extracted entities were normalized to standard ontologies via BioLORD embedding similarity. Phenotypes were mapped to the Human Phenotype Ontology (HPO), common diseases to ICD-10-CM, and rare diseases to Orphanet (ORPHA).

\textbf{Stage-V: manual verification of data quality.} 
To validate the reliability of the automated pipeline, a panel of three senior physicians, with experience ranging from 6 to 10 years, reviewed a set of random sample of 150 cases. The review protocol assessed \textbf{two primary dimensions}: validity, ensuring extracted symptoms and ontology mappings were clinically accurate; and completeness, verifying that no critical diagnostic clues were omitted or leaked. 
As shown in Tab.~\ref{tab:human_verify_results}, the pipeline achieves high fidelity: \textbf{disease label accuracy} reaches 98.7\% ($148/150$), while \textbf{symptom extraction} consistently maintains both precision and recall above 90\%. 
The \textbf{overall accuracy}~(cases free from extraction errors) stands at 95.3\% ($143/150$). Note that, to ensure rigorous benchmarking, the manual validation was extended to the entire test set, guaranteeing high-quality ground-truth for all reported metrics.

\begin{table}[t]
\centering
\caption{\textbf{Manual verification of data quality.} 
Evaluation results from a stratified sample of 150 cases reviewed by senior physicians. ``Overall ACC'' denotes the proportion of cases where both diagnosis and symptom extraction were error-free.}
\label{tab:human_verify_results}
\resizebox{\textwidth}{!}{%
\begin{tabular}{lccccc}
\toprule
\textbf{Metric} & \textbf{MIMIC-Common} & \textbf{PMC-Patients} & \textbf{MIMIC-Rare} & \textbf{RareArena} & \textbf{Xinhua-Rare} \\ \midrule
\textbf{Sample Size} & 25 & 25 & 25 & 25 & 50 \\
\textbf{Overall Accuracy} & 23/25 (92.0\%) & 23/25 (92.0\%) & 25/25 (100.0\%) & 24/25 (96.0\%) & 48/50 (96.0\%) \\
\textbf{Disease Label Accuracy} & 24/25 (96.0\%) & 25/25 (100.0\%) & 25/25 (100.0\%) & 24/25 (96.0\%) & 50/50 (100.0\%) \\
\textbf{Symptom Recall} & 208/218 (95.4\%) & 175/208 (84.1\%) & 254/254 (100.0\%) & 190/210 (90.5\%) & 201/215 (93.5\%) \\
\textbf{Symptom Precision} & 208/219 (95.0\%) & 175/180 (97.2\%) & 254/269 (94.4\%) & 190/195 (97.4\%) & 201/219 (91.8\%) \\
\textbf{Data Integrity (No Leakage)} & 24/25 (96.0\%) & 24/25 (96.0\%) & 22/25 (88.0\%) & 25/25 (100.0\%) & 50/50 (100.0\%) \\ \midrule
\textbf{ICD-10 Accuracy} & 21/25 (84.0\%) & 22/25 (88.0\%) & - & - & - \\
\textbf{ORPHA Accuracy} & - & - & 29/36 (80.6\%) & 22/25 (88.0\%) & 45/50 (90.0\%) \\
\textbf{HPO Accuracy} & 203/219 (92.7\%) & 172/180 (95.6\%) & 251/269 (93.3\%) & 181/195 (92.8\%) & 207/219 (94.5\%) \\ \bottomrule
\end{tabular}%
}
\end{table}

\noindent \textbf{Final Dataset Components and Statistics}

In this section, we start by detailing the retrieval corpus, 
followed by the training and evaluation datasets for model development.

\noindent \textbf{Medical retrieval corpus.}  
The retrieval corpus integrates diverse medical knowledge to mitigate coverage gaps, encompassing both common and rare diseases through three major components:
\textbf{(i) disease information guideline}. This component contains a structured knowledge base to support evidence-based reasoning, including disease-symptom guideline for 16,371 diseases. For rare diseases, we integrated expert-curated phenotype data from Orphanet; for common diseases, we utilized DeepSeek-V3 to summarize symptom profiles from authoritative web sources ({\em e.g.}, Mayo Clinic, NIH). In total, this component contains 257,022 disease–phenotype/symptom pairs~(142,141 common; 114,881 rare), as each disease may correspond to multiple symptoms. Each item is mapped to standard ICD, ORPHA, and HPO terminologies, 
and the dataset achieves complete coverage of ICD codes (to one decimal place) and 38.68\% coverage of ORPHA codes, with over 50\% of HPO terms included;
\textbf{(ii) patient record database}. To enable similar-case retrieval, we constructed a massive repository with 155,442 patient records, with the disease distribution following a long-tailed pattern across 14 major body systems~\cite{Miller2000MEDLINEplusBA}.
This database comprises validated diagnoses, clinical presentations, and medication histories, that met quality standards but were not selected for the training set, ensuring the retrieval corpus retains the diversity of real-world distributions without leaking ground-truth training data. The final database integrates 44,821 cases from MIMIC-IV, 49,633 from PMC-Patients, 6,414 from MedDialog, 46,518 from RareArena, and 324 from RareBench;
\textbf{(iii) clinical knowledge collection}. To provide broad semantic coverage, we incorporated 3.31 million biomedical documents from Wikipedia, 23.9 million PubMed articles, entries into chunks of 1,000 characters to facilitate dense retrieval.

\noindent \textbf{Training \& evaluation dataset.}  
We curated a diverse cohort of 24,142 clinical cases, comprising clinical presentations paired with confirmed diagnoses from seven sources spanning America, Asia, and Europe. 
Following strict quality control for clarity and causality, cases were categorized into common and rare disease groups using the Orphanet coding system. Common diseases constitute 73.1\% of the cohort, drawn from MIMIC-C (7,257), PMC-Patients (6,421), MedDialog (3,206), and Mendeley (757). The remaining 26.9\% represents rare diseases, sourced from RareArena (3,242), MIMIC-R (2,184), RareBench (277) and Xinhua-Rare (798). This collection exhibits noteable phenotypic diversity, averaging 4–12 symptoms per case and covering over 3,000 distinct diseases.
The five \textbf{in-distribution} datasets (MIMIC, PMC-Patients, MedDialog, RareArena, and RareBench) were partitioned at a 3:1 ratio for training and evaluation. 
To assess generalization to unseen data sources, the remaining two datasets, Mendeley~\cite{Shafi2025ASD} and Xinhua-Rare~\cite{Zhao2025AnAS}, were reserved exclusively for \textbf{out-of-distribution} zero-shot evaluation.

\subsection{Agentic RAG Framework}
\label{sec:problem_form}

We model the clinical diagnostic process as a partially observable Markov Decision Process~(MDP) within a reinforcement learning (RL) framework. This system comprises two primary components: 
an \textbf{LLM-based agent} (policy $\mathcal{M}_\theta$), responsible for step-wise decision-making; and an \textbf{external environment} ($\mathcal{E}$), denoting the medical knowledge base~({\em e.g.}, disease guidelines, patient records and other medical knowledge). The objective is therefore to optimize the policy $\mathcal{M}_\theta$ to generate a diagnostic trajectory that maximizes both process validity (correct evidence gathering) and diagnostic accuracy.

\noindent \textbf{Action space.}
To simulate clinical reasoning, we define the action space $\mathcal{A}$ as having both retrieval and reasoning.
The agent can select from five atomic operations:
\textbf{(i) internal processing}. $\langle\text{\texttt{reason}}\rangle$ allows the agent to synthesize current evidence and update hypotheses; $\langle\text{\texttt{diagnose}}\rangle$ is the terminal action where the agent commits to a final diagnosis;
\textbf{(ii) external retrieval}. The agent acts as a query generator for specific tools. $\langle\text{\texttt{lookup}}\rangle$ queries disease guidelines; $\langle\text{\texttt{match}}\rangle$ retrieves similar patient cases based on phenotype lists; and $\langle\text{\texttt{search}}\rangle$ performs free-text queries for general medical knowledge.

\noindent \textbf{Interaction dynamics.}
At each step, the agent is trained to generate a tuple $a_t = (\alpha_t, \tau_t) = \mathcal{M}_{\theta}(S_{t-1})$, where $\alpha_t$ denotes the action type and $\tau_t$ represents the textual content ({\em e.g.}, a search query or reasoning thought),
where $\mathcal{S}_{t-1} = \{\mathcal{S}_0, a_1, f_1, \ldots, a_{t-1}, f_{t-1}\}$, containing the patient's clinical presentation~($\mathcal{S}_0$), {\em e.g.}, symptoms, history, and examination findings, the accumulated trajectory of prior actions and observations until step $t-1$. If the action denotes external retrieval, the environment will then execute the query and return the search results or similar case, {\em i.e.}, $f_t = \mathcal{E}(\alpha_t, \tau_t)$, and the state is then updated to $\mathcal{S}_{t} = \mathcal{S}_{t-1} \cup \{a_t, f_t\}$. 
The process repeats until the terminal $\langle\text{\texttt{diagnose}}\rangle$ action is issued.

\subsection{Reward Mechanism}

To steer the model toward transparent and accurate diagnostics, we designed a composite reward function that balances structural adherence, exploration diversity, evidence quality, and final diagnosis accuracy.

\noindent \textbf{Format alignment reward~($\sigma_f$).}
To ensure the agent strictly follows clinical protocols, we employ a binary gating coefficient, $\sigma_f$. This coefficient is set to 1 only if the output strictly adheres to all formatting constraints (e.g., correct tag pairing, valid iteration structure) and 0 otherwise. Any violation nullifies the reward for the entire trajectory, enforcing rigid adherence to the task specification.

\noindent \textbf{Trajectory exploration reward~($\sigma_{\text{div}}$).}
To prevent the model from collapsing into repetitive, deterministic diagnostic pathways, we penalize over-represented action sequences. We calculate the frequency ratio $r$ of the current trajectory within the training population. The diversity coefficient is defined as:
\begin{align}
\sigma_{\text{div}} = 
\begin{cases}
1-r, & \text{if } r > \tau_{\text{freq}}, \\
1, & \text{otherwise}.
\end{cases}
\end{align}

where $\tau_{\text{freq}}$ is a threshold for allowable repetition.

\noindent \textbf{Evidence acquisition rewards.} 
We incentivize high-quality information gathering through two components: \textbf{(i) patient matching~($\sigma_m$)}. 
This reward encourages the agent to iteratively refine phenotype or symptom queries to find similar past cases. A positive reward ($+0.5$) is granted if a retrieved case matches the ground-truth diagnosis. To promote efficiency, a penalty is applied for each \texttt{<match>} operation ($-0.1$, capped at $-0.3$). Furthermore, to ensure meaningful iteration, $\sigma_m$ is zeroed if the agent fails to vary the phenotype set between consecutive queries~(at least two phenotypes must change);
\textbf{(ii) search relevance~($\sigma_s$).} 
We quantify the relevance of general queries by calculating the token-level overlap between the retrieved disease terms and the ground-truth, hereby encouraging the model to propose correct or relevant candidates that may facilitate the final answer. Let $f_{\text{match}}$ be the fraction of matched tokens; the reward is scaled non-linearly to encourage partial matches early in training: $\sigma_s = \sqrt[3]{f_{\text{match}}}$.

\noindent \textbf{Diagnostic accuracy reward~($\sigma_d$).}
The final component evaluates the correctness of the committed diagnosis, diseases highlighted in the answer (with the special format as \texttt{\textbackslash textbf\{\}} within $\langle\text{\texttt{diagnose}}\rangle ... \langle\text{\texttt{/diagnose}}\rangle$).
We compute a token-level similarity score between the predicted diagnosis and the ground-truth, 
{\em i.e.}, $\text{sim}_{\text{diag}}$. 
This is linearly rescaled to $[0.2, 0.8]$ and adjusted by the accumulated matching reward $\sigma_m$,
which can either increase the reward (for correct matching and reasoning) or decrease it (to penalize excessive or redundant matching and insufficient diversity):
\begin{equation}
\sigma_d = 0.2 + 0.6 \cdot \text{sim}_{\text{diag}} + \sigma_m
\end{equation}
This formulation explicitly links the final reward to the quality of the preceding evidence-gathering process.

\noindent \textbf{Total rewards.}
The final optimization signal $\sigma_{\text{total}}$ is a weighted sum of component rewards, gated by format validity and modulated by diversity:
\begin{equation}
\sigma_{\text{total}} = \text{clip}{[0,1]} \left[ \sigma_f \cdot \sigma{_\text{div}} \cdot (w_m \sigma_m + w_s \sigma_s + w_d \sigma_d) \right]
\end{equation}

\subsection{Training Implementation}
To incorporate the tailored reward mechanism into workflow optimization, we adopt the following training methods regime.

\noindent \textbf{Interleaved agent-environment rollout.}
Unlike standard language model training, our framework requires dynamic interaction during sequence generation. We implemented an inference engine that halts generation upon detecting tool invocation tokens (e.g., $\langle\text{\texttt{lookup}}\rangle$). The system extracts the query, retrieves information from external servers ({\em e.g.}, Wikipedia, PubMed, and medical guidelines), and uses a dedicated summarization model (Qwen-3-235B) to synthesize the retrieved documents. 
This feedback is appended to the context, and generation resumes.

\noindent \textbf{Group Relative Policy Optimization (GRPO).} We optimize the policy using GRPO. For each query $q$, we sample a group of $G$ outputs $\{c_i\}_{i=1}^G$ from the old policy $\mathcal{M}_{\theta_{old}}$. The group-relative advantage is computed by standardizing the rewards within the group: $\hat{A}_{i} = (R_i - \text{mean}(\{R\})) / \text{std}(\{R\})$. The optimization objective is to minimize the following loss function, which incorporates token-level likelihood updates and KL-divergence regularization against a reference model $\mathcal{M}_{\text{ref}}$:
\begin{equation}
\mathcal{L}_{\text{GRPO}}(\theta) = \mathbb{E}_{q, \{c_i\} \sim \mathcal{M}_{\theta_{\text{old}}}} \left[ \frac{1}{G} \sum_{i=1}^G \frac{1}{|c_i|} \sum_{t=1}^{|c_i|} \left( -\hat{A}_{i} \log \mathcal{M}_\theta(c_{i,t}|q, c_{i,<t}) + \beta D_{\mathrm{KL}}(\mathcal{M}_\theta || \mathcal{M}_{\text{ref}}) \right) \right]
\end{equation}

\noindent \textbf{Multi-stage reward adaptation.}
To address the competing objectives of retrieval and diagnosis, we employed a multi-stage training strategy. We observed that optimizing all rewards simultaneously led to suboptimal convergence. Therefore, we sequentially prioritized objectives:
\textbf{stage 1--3:} we individually targeted $\sigma_s$, $\sigma_m$, and $\sigma_d$ in sequence. In each stage, the target component's weight was set to 0.9, while others were dampened to 0.05; \textbf{stage 4:} we balanced the weights ($w_s=0.3, w_m=0.3, w_d=0.4$) for joint optimization. Notably, prioritizing the patient matching reward~($\sigma_m$) in intermediate stages yielded greater improvements in final diagnostic accuracy than focusing solely on the diagnostic outcome, validating the hypothesis that robust evidence gathering is a prerequisite for accurate clinical decision-making.

\newpage
\bibliographystyle{unsrt} 
\bibliography{references} 

@article{wu2023can,
  title={Can gpt-4v (ision) serve medical applications? case studies on gpt-4v for multimodal medical diagnosis},
  author={Wu, Chaoyi and Lei, Jiayu and Zheng, Qiaoyu and Zhao, Weike and Lin, Weixiong and Zhang, Xiaoman and Zhou, Xiao and Zhao, Ziheng and Zhang, Ya and Wang, Yanfeng and others},
  journal={arXiv preprint arXiv:2310.09909},
  year={2023}
}

@article{Asgari2025AFT,
  title={A framework to assess clinical safety and hallucination rates of LLMs for medical text summarisation},
  author={Elham Asgari and Nina Montaña Brown and Magda Dubois and Saleh Khalil and Jasmine Balloch and Joshua Au Yeung and Dominic Pimenta},
  journal={NPJ Digital Medicine},
  year={2025},
  volume={8},
  url={https://api.semanticscholar.org/CorpusID:272600540}
}

@article{Liu2025AGM,
  title={A generalist medical language model for disease diagnosis assistance.},
  author={Xiaohong Liu and Hao Liu and Guoxing Yang and Zeyu Jiang and Shuguang Cui and Zhaoze Zhang and Huan Wang and Liyuan Tao and Yongchang Sun and Zhu Song and Tianpei Hong and Jin Yang and Tianrun Gao and Jiangjiang Zhang and Xiaohu Li and Jing Zhang and Ye Sang and Zhao Yang and Kanmin Xue and Song Wu and Ping Zhang and Jian Yang and Chunli Song and Guangyu Wang},
  journal={Nature medicine},
  year={2025},
  url={https://api.semanticscholar.org/CorpusID:275425003}
}

@article{sandmann2025benchmark,
  title={Benchmark evaluation of DeepSeek large language models in clinical decision-making},
  author={Sandmann, Sarah and Hegselmann, Stefan and Fujarski, Michael and Bickmann, Lucas and Wild, Benjamin and Eils, Roland and Varghese, Julian},
  journal={Nature Medicine},
  pages={1--1},
  year={2025},
  publisher={Nature Publishing Group US New York}
}

@incollection{Tenny2025Evidence,
  author    = {Steven Tenny and Matthew A. Varacallo},
  title     = {Evidence-Based Medicine},
  booktitle = {StatPearls [Internet]},
  publisher = {StatPearls Publishing},
  year      = {2025},
  address   = {Treasure Island (FL)},
  month     = {jan},
  note      = {Updated 2024 Sep 10},
  url       = {https://www.ncbi.nlm.nih.gov/books/NBK470182/},
  pmid      = {29262040}
}

@article{guyatt2002users,
  title={Users' guides to the medical literature: a manual for evidence-based clinical practice},
  author={Guyatt, Gordon and Rennie, Drummond and Satya-Murti, S},
  journal={JAMA-Journal of the American Medical Association-International Edition},
  volume={287},
  number={11},
  pages={1463},
  year={2002},
  publisher={Chicago: American Medical Association, 1960-}
}

@article{guyatt1992evidence,
  title={Evidence-based medicine: a new approach to teaching the practice of medicine},
  author={Guyatt, Gordon and Cairns, John and Churchill, David and Cook, Deborah and Haynes, Brian and Hirsh, Jack and Irvine, Jan and Levine, Mark and Levine, Mitchell and Nishikawa, Jim and others},
  journal={jama},
  volume={268},
  number={17},
  pages={2420--2425},
  year={1992},
  publisher={American Medical Association}
}

@misc{openevidence_home_2026,
  author = {{OpenEvidence}},
  title = {The leading medical information platform},
  year = {2026},
  url = {https://www.openevidence.com/},
  note = {Accessed: 2026-01-26}
}

@misc{anthropic_healthcare_2026,
  author = {{Anthropic}},
  title = {Advancing {Claude} in healthcare and the life sciences},
  year = {2026},
  month = {jan},
  day = {11},
  url = {https://www.anthropic.com/news/healthcare-life-sciences},
  note = {Accessed: 2026-01-26}
}

@misc{openai_chatgpt_health,
  author = {{OpenAI}},
  title = {Introducing {ChatGPT} {Health}},
  year = {2024},
  month = {oct},
  day = {17},
  url = {https://openai.com/index/introducing-chatgpt-health/},
  note = {Accessed: 2026-01-26}
}

@article{qiu2024llm,
  title={LLM-based agentic systems in medicine and healthcare},
  author={Qiu, Jianing and Lam, Kyle and Li, Guohao and Acharya, Amish and Wong, Tien Yin and Darzi, Ara and Yuan, Wu and Topol, Eric J},
  journal={Nature Machine Intelligence},
  volume={6},
  number={12},
  pages={1418--1420},
  year={2024},
  publisher={Nature Publishing Group UK London}
}

@article{zou2025rise,
  title={The rise of agentic AI teammates in medicine},
  author={Zou, James and Topol, Eric J},
  journal={The Lancet},
  volume={405},
  number={10477},
  pages={457},
  year={2025},
  publisher={Elsevier}
}

@article{kresevic2024optimization,
  title={Optimization of hepatological clinical guidelines interpretation by large language models: a retrieval augmented generation-based framework},
  author={Kresevic, Simone and Giuffr{\`e}, Mauro and Ajcevic, Milos and Accardo, Agostino and Croc{\`e}, Lory S and Shung, Dennis L},
  journal={NPJ digital medicine},
  volume={7},
  number={1},
  pages={102},
  year={2024},
  publisher={Nature Publishing Group UK London}
}

@article{leblanc2021rare,
  title={Rare disease patient matchmaking: development and outcomes of an internet case-finding strategy in the Undiagnosed Diseases Network},
  author={LeBlanc, Kimberly and Glanton, Emily and Nagy, Anna and Bater, Jorick and Berro, Tala and McGuinness, Molly A and Studwell, Courtney and Undiagnosed Diseases Network and Might, Matthew},
  journal={Orphanet journal of rare diseases},
  volume={16},
  number={1},
  pages={210},
  year={2021},
  publisher={Springer}
}

@inproceedings{li-etal-2024-mmedagent,
    title = "{MM}ed{A}gent: Learning to Use Medical Tools with Multi-modal Agent",
    author = "Li, Binxu  and
      Yan, Tiankai  and
      Pan, Yuanting  and
      Luo, Jie  and
      Ji, Ruiyang  and
      Ding, Jiayuan  and
      Xu, Zhe  and
      Liu, Shilong  and
      Dong, Haoyu  and
      Lin, Zihao  and
      Wang, Yixin",
    editor = "Al-Onaizan, Yaser  and
      Bansal, Mohit  and
      Chen, Yun-Nung",
    booktitle = "Findings of the Association for Computational Linguistics: EMNLP 2024",
    month = nov,
    year = "2024",
    address = "Miami, Florida, USA",
    publisher = "Association for Computational Linguistics",
    url = "https://aclanthology.org/2024.findings-emnlp.510/",
    doi = "10.18653/v1/2024.findings-emnlp.510",
    pages = "8745--8760"
}

@article{hurst2024gpt,
  title={Gpt-4o system card},
  author={Hurst, Aaron and Lerer, Adam and Goucher, Adam P and Perelman, Adam and Ramesh, Aditya and Clark, Aidan and Ostrow, AJ and Welihinda, Akila and Hayes, Alan and Radford, Alec and others},
  journal={arXiv preprint arXiv:2410.21276},
  year={2024}
}

@article{sellergren2025medgemma,
  title={MedGemma Technical Report},
  author={Sellergren, Andrew and Kazemzadeh, Sahar and Jaroensri, Tiam and Kiraly, Atilla and Traverse, Madeleine and Kohlberger, Timo and Xu, Shawn and Jamil, Fayaz and Hughes, C{\'\i}an and Lau, Charles and others},
  journal={arXiv preprint arXiv:2507.05201},
  year={2025}
}

@article{ding2025evaluation,
  title={Evaluation and practical application of prompt-driven ChatGPTs for EMR generation},
  author={Ding, Hanlin and Xia, Wenjie and Zhou, Yujia and Wei, Lei and Feng, Yipeng and Wang, Zi and Song, Xuming and Li, Rutao and Mao, Qixing and Chen, Bing and others},
  journal={npj Digital Medicine},
  volume={8},
  number={1},
  pages={77},
  year={2025},
  publisher={Nature Publishing Group UK London}
}

@article{wang2024prompt,
  title={Prompt engineering in consistency and reliability with the evidence-based guideline for LLMs},
  author={Wang, Li and Chen, Xi and Deng, XiangWen and Wen, Hao and You, MingKe and Liu, WeiZhi and Li, Qi and Li, Jian},
  journal={NPJ digital medicine},
  volume={7},
  number={1},
  pages={41},
  year={2024},
  publisher={Nature Publishing Group UK London}
}

@article{Johnson2023MIMICIVAF,
  title={MIMIC-IV, a freely accessible electronic health record dataset},
  author={Alistair E. W. Johnson and Lucas Bulgarelli and Lu Shen and Alvin Gayles and Ayad Shammout and Steven Horng and Tom J. Pollard and Benjamin Moody and Brian Gow and Li-wei H. Lehman and Leo Anthony Celi and Roger G. Mark},
  journal={Scientific Data},
  year={2023},
  volume={10},
  url={https://api.semanticscholar.org/CorpusID:255439889}
}

@article{Miller2000MEDLINEplusBA,
  title={MEDLINEplus: building and maintaining the National Library of Medicine's consumer health Web service.},
  author={Naomi Miller and Eve-Marie Lacroix and Joyce Backus},
  journal={Bulletin of the Medical Library Association},
  year={2000},
  volume={88 1},
  pages={
          11-7
        },
  url={https://api.semanticscholar.org/CorpusID:42673613}
}

@article{Yang2024Qwen25TR,
  title={Qwen2.5 Technical Report},
  author={Qwen An Yang and Baosong Yang and Beichen Zhang and Binyuan Hui and Bo Zheng and Bowen Yu and Chengyuan Li and Dayiheng Liu and Fei Huang and Guanting Dong and Haoran Wei and Huan Lin and Jian Yang and Jianhong Tu and Jianwei Zhang and Jianxin Yang and Jiaxin Yang and Jingren Zhou and Junyang Lin and Kai Dang and Keming Lu and Keqin Bao and Kexin Yang and Le Yu and Mei Li and Mingfeng Xue and Pei Zhang and Qin Zhu and Rui Men and Runji Lin and Tianhao Li and Tingyu Xia and Xingzhang Ren and Xuancheng Ren and Yang Fan and Yang Su and Yi-Chao Zhang and Yunyang Wan and Yuqi Liu and Zeyu Cui and Zhenru Zhang and Zihan Qiu and Shanghaoran Quan and Zekun Wang},
  journal={ArXiv},
  year={2024},
  volume={abs/2412.15115},
  url={https://api.semanticscholar.org/CorpusID:274859421}
}

@article{dubey2024llama,
  title={The llama 3 herd of models},
  author={Dubey, Abhimanyu and Jauhri, Abhinav and Pandey, Abhinav and Kadian, Abhishek and Al-Dahle, Ahmad and Letman, Aiesha and Mathur, Akhil and Schelten, Alan and Yang, Amy and Fan, Angela and others},
  journal={arXiv e-prints},
  pages={arXiv--2407},
  year={2024}
}

@article{Jin2023BioCPTCP,
  title={BioCPT: Contrastive Pre-trained Transformers with Large-scale PubMed Search Logs for Zero-shot Biomedical Information Retrieval},
  author={Qiao Jin and Won Kim and Qingyu Chen and Donald C. Comeau and Lana Yeganova and John Wilbur and Zhiyong Lu},
  journal={Bioinformatics},
  year={2023},
  volume={39 11},
  url={https://api.semanticscholar.org/CorpusID:259316759}
}

@article{hager2024evaluation,
  title={Evaluation and mitigation of the limitations of large language models in clinical decision-making},
  author={Hager, Paul and Jungmann, Friederike and Holland, Robbie and Bhagat, Kunal and Hubrecht, Inga and Knauer, Manuel and Vielhauer, Jakob and Makowski, Marcus and Braren, Rickmer and Kaissis, Georgios and others},
  journal={Nature medicine},
  volume={30},
  number={9},
  pages={2613--2622},
  year={2024},
  publisher={Nature Publishing Group US New York}
}

@article{tu2024towards,
  title={Towards conversational diagnostic AI},
  author={Tu, Tao and Palepu, Anil and Schaekermann, Mike and Saab, Khaled and Freyberg, Jan and Tanno, Ryutaro and Wang, Amy and Li, Brenna and Amin, Mohamed and Tomasev, Nenad and others},
  journal={arXiv preprint arXiv:2401.05654},
  year={2024}
}

@article{qiu2025evolving,
  title={Evolving Diagnostic Agents in a Virtual Clinical Environment},
  author={Qiu, Pengcheng and Wu, Chaoyi and Liu, Junwei and Zheng, Qiaoyu and Liao, Yusheng and Wang, Haowen and Yue, Yun and Fan, Qianrui and Zhen, Shuai and Wang, Jian and others},
  journal={arXiv preprint arXiv:2510.24654},
  year={2025}
}

@article{liao2025ehr,
  title={EHR-R1: A Reasoning-Enhanced Foundational Language Model for Electronic Health Record Analysis},
  author={Liao, Yusheng and Wu, Chaoyi and Liu, Junwei and Jiang, Shuyang and Qiu, Pengcheng and Wang, Haowen and Yue, Yun and Zhen, Shuai and Wang, Jian and Fan, Qianrui and others},
  journal={arXiv preprint arXiv:2510.25628},
  year={2025}
}

@article{xia2024mmed,
  title={Mmed-rag: Versatile multimodal rag system for medical vision language models},
  author={Xia, Peng and Zhu, Kangyu and Li, Haoran and Wang, Tianze and Shi, Weijia and Wang, Sheng and Zhang, Linjun and Zou, James and Yao, Huaxiu},
  journal={arXiv preprint arXiv:2410.13085},
  year={2024}
}

@article{wu2024medical,
  title={Medical graph rag: Towards safe medical large language model via graph retrieval-augmented generation},
  author={Wu, Junde and Zhu, Jiayuan and Qi, Yunli and Chen, Jingkun and Xu, Min and Menolascina, Filippo and Grau, Vicente},
  journal={arXiv preprint arXiv:2408.04187},
  year={2024}
}

@article{singh2025agentic,
  title={Agentic retrieval-augmented generation: A survey on agentic rag},
  author={Singh, Aditi and Ehtesham, Abul and Kumar, Saket and Khoei, Tala Talaei},
  journal={arXiv preprint arXiv:2501.09136},
  year={2025}
}

@inproceedings{xiong2024benchmarking,
  title={Benchmarking retrieval-augmented generation for medicine},
  author={Xiong, Guangzhi and Jin, Qiao and Lu, Zhiyong and Zhang, Aidong},
  booktitle={Findings of the Association for Computational Linguistics ACL 2024},
  pages={6233--6251},
  year={2024}
}

@article{gao2023retrieval,
  title={Retrieval-augmented generation for large language models: A survey},
  author={Gao, Yunfan and Xiong, Yun and Gao, Xinyu and Jia, Kangxiang and Pan, Jinliu and Bi, Yuxi and Dai, Yixin and Sun, Jiawei and Wang, Haofen and Wang, Haofen},
  journal={arXiv preprint arXiv:2312.10997},
  volume={2},
  number={1},
  year={2023}
}

@article{Pham2024TowardsRM,
  title={Towards Reliable Medical Question Answering: Techniques and Challenges in Mitigating Hallucinations in Language Models},
  author={Duy Khoa Pham and Quoc Bao Vo},
  journal={ArXiv},
  year={2024},
  volume={abs/2408.13808},
  url={https://api.semanticscholar.org/CorpusID:271957051}
}

@article{wu2025towards,
  title={Towards evaluating and building versatile large language models for medicine},
  author={Wu, Chaoyi and Qiu, Pengcheng and Liu, Jinxin and Gu, Hongfei and Li, Na and Zhang, Ya and Wang, Yanfeng and Xie, Weidi},
  journal={npj Digital Medicine},
  volume={8},
  number={1},
  pages={58},
  year={2025},
  publisher={Nature Publishing Group UK London}
}

@article{johri2025evaluation,
  title={An evaluation framework for clinical use of large language models in patient interaction tasks},
  author={Johri, Shreya and Jeong, Jaehwan and Tran, Benjamin A and Schlessinger, Daniel I and Wongvibulsin, Shannon and Barnes, Leandra A and Zhou, Hong-Yu and Cai, Zhuo Ran and Van Allen, Eliezer M and Kim, David and others},
  journal={Nature medicine},
  volume={31},
  number={1},
  pages={77--86},
  year={2025},
  publisher={Nature Publishing Group US New York}
}

@article{Xu2024BMRetrieverTL,
  title={BMRetriever: Tuning Large Language Models as Better Biomedical Text Retrievers},
  author={Ran Xu and Wenqi Shi and Yue Yu and Yuchen Zhuang and Yanqiao Zhu and May Dongmei Wang and Joyce C. Ho and Chao Zhang and Carl Yang},
  journal={ArXiv},
  year={2024},
  volume={abs/2404.18443},
  url={https://api.semanticscholar.org/CorpusID:269448997}
}

@article{Delile2024GraphBasedRC,
  title={Graph-Based Retriever Captures the Long Tail of Biomedical Knowledge},
  author={Julien Delile and Srayanta Mukherjee and Anton Van Pamel and Leonid Zhukov},
  journal={ArXiv},
  year={2024},
  volume={abs/2402.12352},
  url={https://api.semanticscholar.org/CorpusID:267751251}
}

@article{wu2024pmc,
  title={PMC-LLaMA: toward building open-source language models for medicine},
  author={Wu, Chaoyi and Lin, Weixiong and Zhang, Xiaoman and Zhang, Ya and Xie, Weidi and Wang, Yanfeng},
  journal={Journal of the American Medical Informatics Association},
  volume={31},
  number={9},
  pages={1833--1843},
  year={2024},
  publisher={Oxford Academic}
}

@article{Zhao2022PMCPatientsAL,
  title={PMC-Patients: A Large-scale Dataset of Patient Notes and Relations Extracted from Case Reports in PubMed Central},
  author={Zhengyun Zhao and Qiao Jin and Sheng Yu},
  journal={ArXiv},
  year={2022},
  volume={abs/2202.13876},
  url={https://api.semanticscholar.org/CorpusID:247158336}
}

@article{Chen2020MedDialogAL,
  title={MedDialog: A Large-scale Medical Dialogue Dataset},
  author={Shu Chen and Zeqian Ju and Xiangyu Dong and Hongchao Fang and Sicheng Wang and Yue Yang and Jiaqi Zeng and Ruisi Zhang and Ruoyu Zhang and Meng Zhou and Penghui Zhu and Pengtao Xie},
  journal={ArXiv},
  year={2020},
  volume={abs/2004.03329},
  url={https://api.semanticscholar.org/CorpusID:215238980}
}

@article{Chen2024RareBenchCL,
  title={RareBench: Can LLMs Serve as Rare Diseases Specialists?},
  author={Xuanzhong Chen and Xiaohao Mao and Qihan Guo and Lun Wang and Shuyang Zhang and Ting Chen},
  journal={Proceedings of the 30th ACM SIGKDD Conference on Knowledge Discovery and Data Mining},
  year={2024},
  url={https://api.semanticscholar.org/CorpusID:267617076}
}

@article{Shafi2025ASD,
  title={A Structured Dataset of Disease-Symptom Associations to Improve Diagnostic Accuracy},
  author={Abdullah Al Shafi and Rowzatul Zannat and Abdul Muntakim and Mahmudul Hasan},
  journal={ArXiv},
  year={2025},
  volume={abs/2506.13610},
  url={https://api.semanticscholar.org/CorpusID:279402210}
}

@article{Zhao2025AnAS,
  title={An Agentic System for Rare Disease Diagnosis with Traceable Reasoning},
  author={Weike Zhao and Chaoyi Wu and Yanjie Fan and Xiaoman Zhang and Pengcheng Qiu and Yuze Sun and Xiao Zhou and Yanfeng Wang and Ya Zhang and Yongguo Yu and Kun Sun and Weidi Xie},
  journal={ArXiv},
  year={2025},
  volume={abs/2506.20430},
  url={https://api.semanticscholar.org/CorpusID:280011877}
}

@article{Chen2023MEDITRON70BSM,
  title={MEDITRON-70B: Scaling Medical Pretraining for Large Language Models},
  author={Zeming Chen and Alejandro Hern'andez Cano and Angelika Romanou and Antoine Bonnet and Kyle Matoba and Francesco Salvi and Matteo Pagliardini and Simin Fan and Andreas Kopf and Amirkeivan Mohtashami and Alexandre Sallinen and Alireza Sakhaeirad and Vinitra Swamy and Igor Krawczuk and Deniz Bayazit and Axel Marmet and Syrielle Montariol and Mary-Anne Hartley and Martin Jaggi and Antoine Bosselut},
  journal={ArXiv},
  year={2023},
  volume={abs/2311.16079},
  url={https://api.semanticscholar.org/CorpusID:265456229}
}

@article{singhal2023large,
  title={Large language models encode clinical knowledge},
  author={Singhal, Karan and Azizi, Shekoofeh and Tu, Tao and Mahdavi, S Sara and Wei, Jason and Chung, Hyung Won and Scales, Nathan and Tanwani, Ajay and Cole-Lewis, Heather and Pfohl, Stephen and others},
  journal={Nature},
  volume={620},
  number={7972},
  pages={172--180},
  year={2023},
  publisher={Nature Publishing Group}
}

@article{qiu2025quantifying,
  title={Quantifying the reasoning abilities of LLMs on clinical cases},
  author={Qiu, Pengcheng and Wu, Chaoyi and Liu, Shuyu and Fan, Yanjie and Zhao, Weike and Chen, Zhuoxia and Gu, Hongfei and Peng, Chuanjin and Zhang, Ya and Wang, Yanfeng and others},
  journal={Nature Communications},
  volume={16},
  number={1},
  pages={9799},
  year={2025},
  publisher={Nature Publishing Group UK London}
}

@article{chen2025enhancing,
  title={Enhancing diagnostic capability with multi-agents conversational large language models},
  author={Chen, Xi and Yi, Huahui and You, Mingke and Liu, WeiZhi and Wang, Li and Li, Hairui and Zhang, Xue and Guo, Yingman and Fan, Lei and Chen, Gang and others},
  journal={NPJ digital medicine},
  volume={8},
  number={1},
  pages={159},
  year={2025},
  publisher={Nature Publishing Group UK London}
}

@inproceedings{chen2025cod,
  title={Cod, towards an interpretable medical agent using chain of diagnosis},
  author={Chen, Junying and Gui, Chi and Gao, Anningzhe and Ji, Ke and Wang, Xidong and Wan, Xiang and Wang, Benyou},
  booktitle={Findings of the Association for Computational Linguistics: ACL 2025},
  pages={14345--14368},
  year={2025}
}

@article{dou2025baichuan,
  title={Baichuan-m2: Scaling medical capability with large verifier system},
  author={Dou, Chengfeng and Liu, Chong and Yang, Fan and Li, Fei and Jia, Jiyuan and Chen, Mingyang and Ju, Qiang and Wang, Shuai and Dang, Shunya and Li, Tianpeng and others},
  journal={arXiv preprint arXiv:2509.02208},
  year={2025}
}

@article{chen2023meditron,
  title={Meditron-70b: Scaling medical pretraining for large language models},
  author={Chen, Zeming and Cano, Alejandro Hern{\'a}ndez and Romanou, Angelika and Bonnet, Antoine and Matoba, Kyle and Salvi, Francesco and Pagliardini, Matteo and Fan, Simin and K{\"o}pf, Andreas and Mohtashami, Amirkeivan and others},
  journal={arXiv preprint arXiv:2311.16079},
  year={2023}
}

@article{feng2025doctoragent,
  title={DoctorAgent-RL: A Multi-Agent Collaborative Reinforcement Learning System for Multi-Turn Clinical Dialogue},
  author={Feng, Yichun and Wang, Jiawei and Zhou, Lu and Li, Yixue},
  journal={arXiv preprint arXiv:2505.19630},
  year={2025}
}

@article{croxford2025evaluating,
  title={Evaluating clinical AI summaries with large language models as judges},
  author={Croxford, Emma and Gao, Yanjun and First, Elliot and Pellegrino, Nicholas and Schnier, Miranda and Caskey, John and Oguss, Madeline and Wills, Graham and Chen, Guanhua and Dligach, Dmitriy and others},
  journal={npj Digital Medicine},
  volume={8},
  number={1},
  pages={640},
  year={2025},
  publisher={Nature Publishing Group UK London}
}

@article{guo2025deepseek,
  title={DeepSeek-R1 incentivizes reasoning in LLMs through reinforcement learning},
  author={Guo, Daya and Yang, Dejian and Zhang, Haowei and Song, Junxiao and Wang, Peiyi and Zhu, Qihao and Xu, Runxin and Zhang, Ruoyu and Ma, Shirong and Bi, Xiao and others},
  journal={Nature},
  volume={645},
  number={8081},
  pages={633--638},
  year={2025},
  publisher={Nature Publishing Group UK London}
}

@article{piran2024disentanglement,
  title={Disentanglement of single-cell data with biolord},
  author={Piran, Zoe and Cohen, Niv and Hoshen, Yedid and Nitzan, Mor},
  journal={Nature Biotechnology},
  volume={42},
  number={11},
  pages={1678--1683},
  year={2024},
  publisher={Nature Publishing Group US New York}
}

\clearpage
\section{Extended Data}
\definecolor{bluefill}{HTML}{E7F4FA}
\definecolor{orangefill}{HTML}{FDEEE0}
\setcounter{figure}{0} 
    \setcounter{table}{0}  
    
    \renewcommand{\figurename}{Extended Data Figure}
    \renewcommand{\tablename}{Extended Data Table}
    
    \renewcommand{\cref}[1]{\Cref{#1}} 
    \crefname{figure}{Extended Data Figure}{Extended Data Figures}
    \crefname{table}{Extended Data Table}{Extended Data Tables}
    \Crefname{figure}{Extended Data Figure}{Extended Data Figures}
    \Crefname{table}{Extended Data Table}{Extended Data Tables}

\begin{figure}[!htb]
    \centering
    \includegraphics[width=1\linewidth]{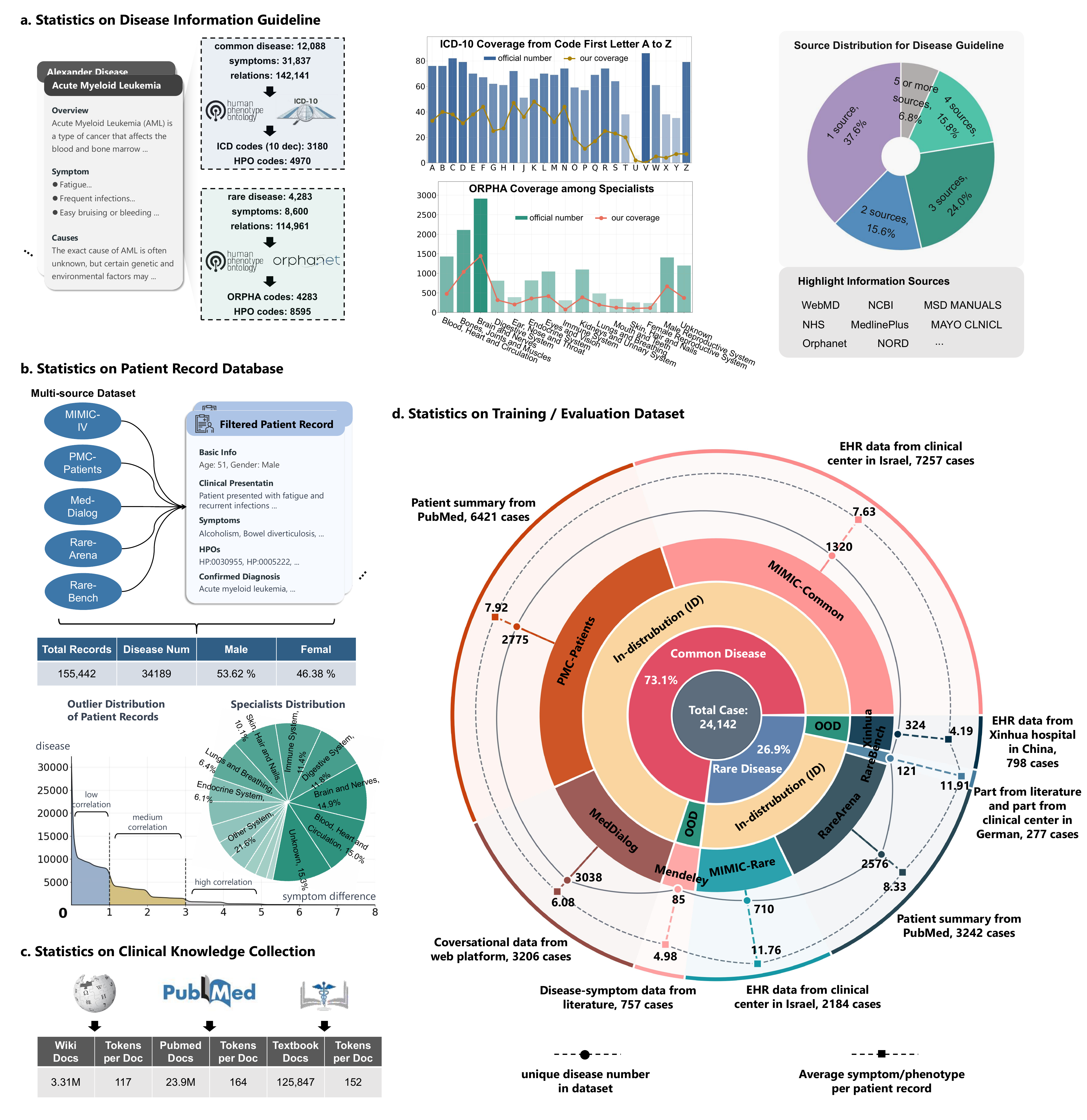}
    \caption{\textbf{Data statistics.} 
    \textbf{a.} Left: Overview of items and their relationships in the disease guideline. Middle: ICD coverage for common diseases and Orpha coverage for rare diseases. Right: Distribution of disease information sources, highlighting major public resources. 
    \textbf{b.} Top: Summary statistics of patient records. Bottom: Distribution of outliers, illustrating discrepancies between real patient disease-symptom associations and guideline expectations; Breakdown of confirmed patient diagnoses by specialty. 
    \textbf{c.} Summary statistics of the clinical knowledge collection. 
    \textbf{d.} Detailed statistics of the seven-center datasets used for training and evaluation.}
    \label{fig:datastatistics_extend}
\end{figure}

\begin{figure}[!h]
    \centering
    \includegraphics[width=1\linewidth]{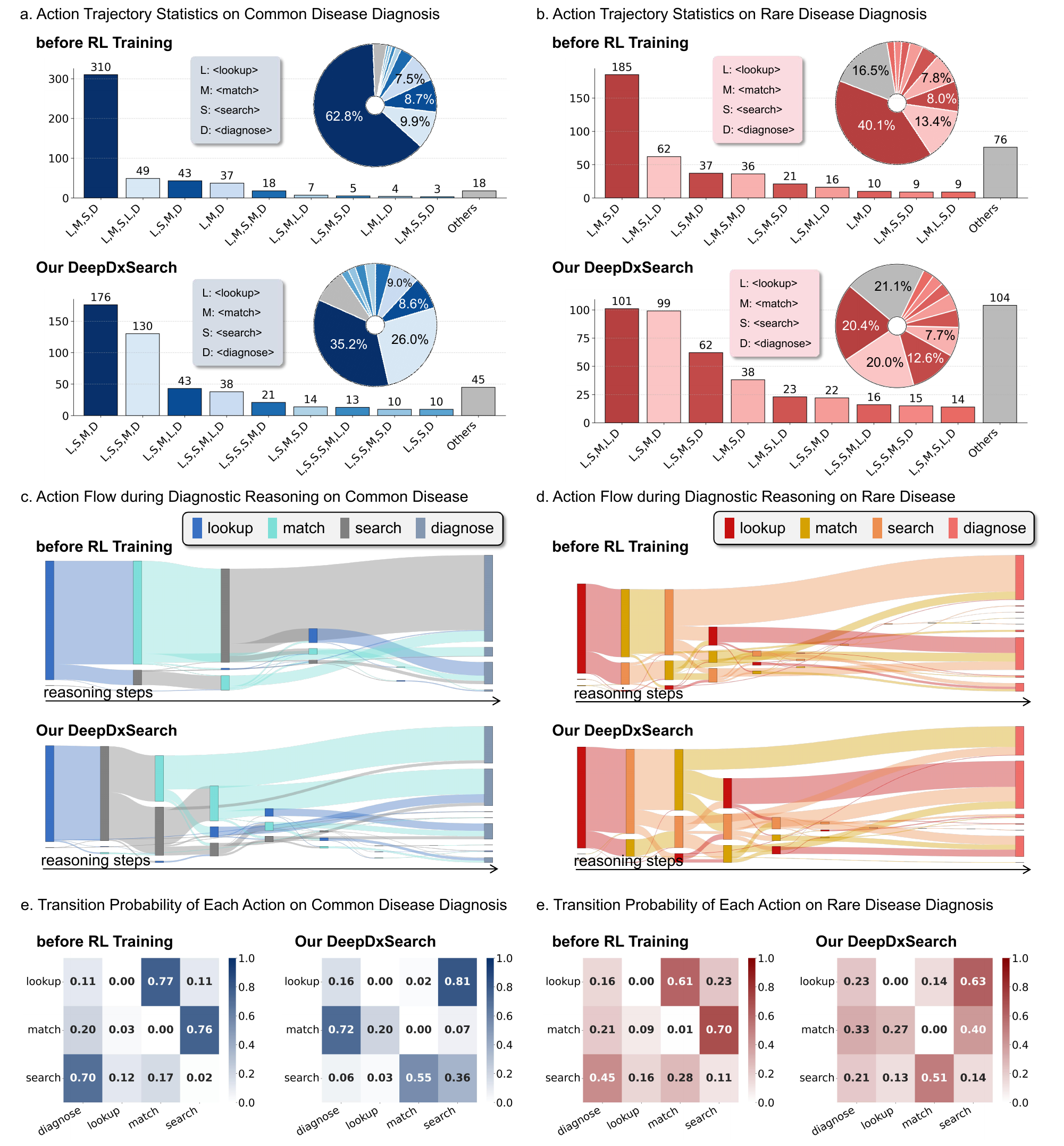}
    \caption{Analysis of reasoning dynamics before and after Deep-DxSearch training. 
    \textbf{a, b.} Statistical comparison of action trajectories for common and rare disease diagnosis. The baseline model (top) exhibits significant algorithmic rigidity, with the fixed sequence ``L,M,S,D'' ($\texttt{<lookup>} \rightarrow \texttt{<match>} \rightarrow \texttt{<search>} \rightarrow \texttt{<diagnose>}$) accounting for 62.8\% and 40.1\% of cases, respectively. In contrast, Deep-DxSearch (bottom) demonstrates increased trajectory diversity (e.g., unique trajectory types increased from 22 to 37 for common diseases), indicating a shift from repetitive heuristics to customized diagnostic paths.
    \textbf{c, d.} Visualization of action flows illustrating the logical progression. The flows confirm a transition from linear, deterministic execution to adaptive, branching investigation suited to case complexity.
    \textbf{e, f.} Action-to-action transition probability matrices. Post-training matrices reveal the emergence of recursive behaviors (e.g., $\texttt{<search>} \rightarrow \texttt{<search>}$ for self-correction) and a smoothed probability distribution in rare diseases, suggesting that the agent's decisions are conditioned on evolving context rather than fixed rules.}
    \label{fig:transition_extend}
\end{figure}

\begin{table}[h]
\centering
\scriptsize 
\setlength{\tabcolsep}{4pt} 
\renewcommand{\arraystretch}{1.25}
\caption{\textbf{Impact of embedding-based de-duplication on diagnostic accuracy.} 
We conducted a rigorous ablation study by filtering cases from the retrieval corpus that exceeded specific cosine similarity thresholds ($\tau$) relative to the query case, utilizing both BioLORD and OpenAI embeddings to measure similarity. \textbf{Rem. \%} denotes the percentage of the retrieval corpus retained after filtering. 
\textbf{Observations:} 
(1) The removal of highly similar cases ($\tau \geq 0.90$) impacts $<2\%$ of the corpus and causes negligible performance variance. 
(2) Diagnostic accuracy remains robust even at moderate thresholds ($\tau=0.80$), where the top $\sim$13--17\% of similar cases are excluded; this confirms that the model's performance drives from synthesizing evidence across a distribution of relevant cases rather than memorizing a specific ``near-duplicate.'' 
(3) A significant performance decline is only observed at aggressive thresholds ($\tau < 0.75$) where the corpus is severely decimated (retention falls below 20\%), indicating the loss of essential medical evidence rather than the removal of leakage.}
\label{tab:embedding_ablation_extend}
\begin{tabular}{c|c|cccccc|c|cccccc}
\toprule
\multirow{3}{*}{\textbf{Thres.}} & \multicolumn{7}{c|}{\textbf{Common Disease Diagnosis}} & \multicolumn{7}{c}{\textbf{Rare Disease Diagnosis}} \\ \cmidrule{2-15} 
 & \textbf{Rem.} & \multicolumn{2}{c}{\textbf{MIMIC-C}} & \multicolumn{2}{c}{\textbf{PMC-Pat.}} & \multicolumn{2}{c|}{\textbf{MedDialog}} & \textbf{Rem.} & \multicolumn{2}{c}{\textbf{MIMIC-R}} & \multicolumn{2}{c}{\textbf{RareArena}} & \multicolumn{2}{c}{\textbf{RareBench}} \\
 & \textbf{\%} & Acc@1 & Acc@5 & Acc@1 & Acc@5 & Acc@1 & Acc@5 & \textbf{\%} & Acc@1 & Acc@5 & Acc@1 & Acc@5 & Acc@1 & Acc@5 \\ \midrule
\multicolumn{15}{l}{\cellcolor{bluefill}\textbf{BioLORD Embedding}} \\ \midrule
1.00 & 100.0 & 31.03 & 41.62 & 43.51 & 52.16 & 47.27 & 58.44 & 100.0 & 34.86 & 46.69 & 31.93 & 41.20 & 71.44 & 81.22 \\
0.95 & 99.9 & 31.16 & 41.71 & 43.33 & 51.80 & 47.27 & 59.77 & 99.2 & 34.43 & 46.32 & 30.85 & 40.79 & 67.59 & 78.14 \\
0.90 & 99.0 & 30.92 & 42.39 & 42.84 & 51.66 & 46.55 & 59.39 & 98.8 & 33.01 & 44.77 & 30.36 & 41.15 & 64.31 & 74.38 \\
0.85 & 94.8 & 30.03 & 39.97 & 41.11 & 50.86 & 45.26 & 58.10 & 97.0 & 32.03 & 44.59 & 29.16 & 39.42 & 61.27 & 71.55 \\
0.80 & 83.1 & 28.43 & 37.51 & 40.32 & 49.17 & 44.59 & 57.48 & 90.9 & 29.45 & 40.26 & 27.41 & 36.03 & 60.17 & 70.82 \\
0.75 & 63.1 & 25.47 & 32.14 & 36.82 & 46.24 & 41.92 & 54.33 & 73.6 & 26.72 & 37.65 & 26.18 & 34.99 & 59.46 & 69.20 \\
0.70 & 39.4 & 23.88 & 31.70 & 32.67 & 44.51 & 39.52 & 50.96 & 44.0 & 22.60 & 32.32 & 23.42 & 33.16 & 56.27 & 65.18 \\ \midrule
\multicolumn{15}{l}{\cellcolor{orangefill}\textbf{OpenAI Embedding}} \\ \midrule
1.00 & 100.0 & 31.03 & 41.62 & 43.51 & 52.16 & 47.27 & 58.44 & 100.0 & 34.86 & 46.69 & 31.93 & 41.20 & 71.44 & 81.22 \\
0.95 & 99.2 & 30.88 & 41.79 & 43.25 & 51.71 & 46.75 & 57.98 & 100.0 & 34.39 & 64.41 & 31.17 & 39.03 & 70.42 & 79.96 \\
0.90 & 98.8 & 30.10 & 41.68 & 43.12 & 51.58 & 46.60 & 57.82 & 99.8 & 34.95 & 64.43 & 31.01 & 39.05 & 67.25 & 75.71 \\
0.85 & 96.9 & 29.85 & 41.35 & 42.75 & 41.15 & 46.15 & 57.30 & 97.7 & 34.30 & 63.85 & 31.65 & 38.67 & 64.42 & 73.85 \\
0.80 & 87.2 & 28.60 & 39.92 & 41.45 & 49.83 & 44.71 & 55.85 & 84.8 & 30.61 & 59.75 & 27.15 & 35.45 & 60.12 & 70.53 \\
0.75 & 55.2 & 25.80 & 36.24 & 38.22 & 47.58 & 40.35 & 50.60 & 51.3 & 24.17 & 48.29 & 24.86 & 31.62 & 54.43 & 64.27 \\
0.70 & 19.8 & 19.45 & 28.20 & 29.19 & 38.80 & 32.54 & 41.15 & 19.2 & 15.80 & 42.60 & 20.20 & 25.80 & 48.10 & 58.50 \\ \bottomrule
\end{tabular}
\end{table}

\begin{table}[h]
\centering
\scriptsize
\caption{\textbf{Action Space Ablation: Performance vs. Efficiency Trade-off.} We report the latency \textit{after} removing each component and the specific time reduction (in parentheses). Performance impact is measured by the cumulative drop in Top-1 Accuracy on Rare Diseases compared to the Full Pipeline.}
\label{tab:action_ablation_extend}
\resizebox{\columnwidth}{!}{%
\begin{tabular}{l p{4.0cm} c c p{3.8cm}}
\toprule
\textbf{System State} & \textbf{Component Removed} & \textbf{Acc@1 Drop} & \textbf{Time Cost (Reduction)} & \textbf{Efficiency Analysis} \\ 
\midrule
\textbf{Full Pipeline} & None & - & 31.78s (-) & Optimal accuracy baseline. \\
\midrule
\textbf{w/o \texttt{<search>}} & Knowledge Searcher \& Summarizer & -8.33\%$^{\dagger}$ & 19.23s (\textbf{-12.55s}) & \textbf{High Cost:} Processing unstructured text is time-consuming but vital for robustness. \\
\midrule
\textbf{w/o \texttt{<lookup>}} & Disease Guideline & -1.88\% & 16.37s (\textbf{-2.86s}) & \textbf{Low Cost:} Fast execution; structures the search space. \\
\midrule
\textbf{w/o \texttt{<match>}} & Patient Record Database & \textbf{-17.46\%} & 10.15s (\textbf{-6.22s}) & \textbf{High Value:} Critical for accuracy; offers best ROI on latency. \\
\midrule
\textbf{w/o \texttt{<reason>}} & Policy Reward & \textbf{-22.14\%} & 10.15s (Base) & \textbf{Foundation:} Core LLM inference without tool support. \\
\bottomrule
\end{tabular}%
}
\vspace{0.1cm}
\raggedright
\footnotesize{$^{\dagger}$ Combined drop from removing the Document Summarizer (-5.61\%) and Knowledge Searcher (-2.72\%).}
\end{table}

\clearpage
\appendix
\section{Supplementary}

\setcounter{figure}{0} 
    \setcounter{table}{0}  
    
    \renewcommand{\figurename}{Supplementary Figure}
    \renewcommand{\tablename}{Supplementary Table}
    
    \renewcommand{\cref}[1]{\Cref{#1}} 
    \crefname{figure}{Supplementary Figure}{Supplementary Figures}
    \crefname{table}{Supplementary Table}{Supplementary Tables}
    \Crefname{figure}{Supplementary Figure}{Supplementary Figures}
    \Crefname{table}{Supplementary Table}{Supplementary Tables}

\subsection{Framework Instruction}
\label{inital_prompt}

\begin{tcolorbox}[
    colback=orange!5!white,
    colframe=orange!80!black,
    width=0.98\textwidth,
    sharp corners=south,
    fontupper=\footnotesize,
    title={System Prompt}
]

You are an AI assistant specializing in diagnosing diseases based on phenotypes or symptoms.

\textbf{Task Description:} \\
Your task is to analyze patient clinical presentation including phenotypes or symptoms and make a final disease diagnosis through systematic medical reasoning using the available tools.

\textbf{Available Tools:}
\begin{enumerate}
    \item \textbf{Disease Information Guideline Lookup Tool:} Use the \texttt{<lookup>} tag to query typical phenotypes or symptoms of specific diseases. \\
    Format: \texttt{<lookup> disease1, disease2... </lookup>} \\
    The system returns common phenotypes for each disease enclosed in a \texttt{<guide>} tag.
    
    \item \textbf{Patient Record Database Match Tool:} Use the \texttt{<match>} tag to submit a list of phenotypes. The system returns similar known cases, including diseases and their corresponding symptoms, enclosed in a \texttt{<refer>} tag. \\
    Format: \texttt{<match> phenotype1, phenotype2, phenotype3... </match>}
    
    \item \textbf{Medical Knowledge Corpus Search Tool:} Use the \texttt{<search>} tag to retrieve knowledge from Wikipedia, PMC, or textbooks using free-text queries (do not use commas within each question). \\
    Format: \texttt{<search> |WIKI| query1, query2... </search>} or \texttt{<search> |PMC| query1, query2... </search>} or \texttt{<search> |BOOK| query1, query2... </search>} \\
    Specify the source using the prefix \texttt{|WIKI|}, \texttt{|PMC|}, or \texttt{|BOOK|}. The system returns the retrieved content in a \texttt{<result>} tag.
\end{enumerate}

\textbf{Allowed Actions:}
\begin{enumerate}
    \item \texttt{<reason> </reason>}: Active action. Use for the analysis process or reasoning chain between actions.
    \item \texttt{<lookup> </lookup>}: Active action. Use to look up up to 10 diseases within one \texttt{<lookup>} tag.
    \item \texttt{<guide> </guide>}: Passive action. Returned by the system after a \texttt{<lookup>} action.
    \item \texttt{<match> </match>}: Active action. Use to match a series of patient cases related to the query phenotypes.
    \item \texttt{<refer> </refer>}: Passive action. Returned by the system after a \texttt{<match>} action.
    \item \texttt{<search> </search>}: Active action. Use to search knowledge from only one source, with up to three queries (separated by commas) per \texttt{<search>} tag.
    \item \texttt{<result> </result>}: Passive action. Returned by the system after a \texttt{<search>} action.
    \item \texttt{<diagnose> </diagnose>}: Active action. Analyze all reference information and synthesize to make the final disease diagnosis.
\end{enumerate}

\textbf{Format Requirements:}
\begin{itemize}
    \item \texttt{<reason>} must appear between two active actions.
    \item \texttt{<lookup>} may appear at most once. The content should only include diseases, not symptoms or phenotypes.
    \item \texttt{<match>} may appear up to three times. The content should only include symptoms or phenotypes, not diseases.
    \item \texttt{<search>} may appear at most twice. The content must follow the \texttt{|Source| query1, query2} format, with up to three queries at a time.
    \item The \texttt{<diagnose>} tag is mandatory at the end. Provide up to five possible disease diagnoses, enclosed in LaTeX bold format: \texttt{\textbackslash textbf\{Disease1\}}, \texttt{\textbackslash textbf\{Disease2\}}, etc.
    \item No text may appear outside of the specified tags.
\end{itemize}

\textbf{Phenotype Query Refinement Guide:} \\
If repeating the \texttt{<match>} step for more patient case references, refine the query phenotypes by one or more of the following:
\begin{itemize}
    \item Adding related phenotypes commonly seen in suspected disease categories
    \item Replacing phenotypes with alternative medical terminology
    \item Including potential complications or associated features
    \item Adding earlier or later stage manifestations
    \item Using symptoms from retrieved cases as references
\end{itemize}

\textbf{Diagnostic Workflow:} \\
The diagnostic workflow is flexible. There is no fixed order for using the \texttt{<lookup>}, \texttt{<match>}, or \texttt{<search>} tools; use them as appropriate. Ensure your disease diagnoses are enclosed with \texttt{\textbackslash textbf\{\}} within the \texttt{<diagnose>} tag, with a maximum of five diagnoses.

\end{tcolorbox}

\subsection{Case Demonstration}

To further clarify the problem formulation, 
we illustrate an exact case about a 58-year-old man with a 6-month history of progressive muscle weakness and neurological symptoms, and is ultimately confirmed as \textbf{Systemic Lupus Erythematosus~(SLE)}. The case was collected from RareArena, accompanied by the traceable diagnostic reasoning generated by \ModelName{}:


\begin{tcolorbox}[
colback=softblue!5!white, 
colframe=softblue!40!white, 
width=0.98\textwidth,
sharp corners=south,
breakable,
fontupper=\footnotesize,
title={Case Illustration: Diagnostic Process of A Patient with Complex Neuromuscular Disorder}
]

\vspace{3pt} \textbf{Clinical Presentation Input:}
A 58-year-old male presents with a 6-month history of progressive muscle weakness and neurological symptoms. He initially noticed double vision and drooping of his left eyelid (ptosis) which gradually worsened. Over the following months, he developed numbness and tingling sensations in both hands and feet, along with profound fatigue that significantly impacted his daily activities. The muscle weakness progressed from proximal muscles to involve distal extremities, eventually rendering him unable to walk independently. He has experienced unintentional weight loss of 15 pounds and a persistent chronic cough. Physical examination reveals facial weakness, pronounced proximal muscle weakness affecting shoulders and hips, muscle atrophy in the extremities, diminished deep tendon reflexes, and distal sensory impairment to vibration and position sense. Laboratory findings show proteinuria on urinalysis and elevated erythrocyte sedimentation rate. Lumbar puncture demonstrates elevated protein levels in cerebrospinal fluid. Muscle biopsy reveals inflammatory myopathy with endomysial fibrosis, regenerating muscle fibers, and scattered ragged red fibers on modified Gomori trichrome stain...

\vspace{10pt}
\noindent\makebox[\linewidth]{\rule{0.9\textwidth}{0.4pt}}
\vspace{10pt}

\textbf{Traceable Diagnostic Reasoning Output of \ModelName{}:}

\vspace{3pt} \textcolor{reasoncolor}{\texttt{<reason>}} The patient presents with a combination of cranial and peripheral neuropathies, muscle weakness, and other systemic symptoms. Key symptoms include ptosis, numbness and tingling, fatigue, and proximal muscle weakness. There are also abnormalities in reflexes, proteinuria, and elevated protein in cerebrospinal fluid. The differential to consider includes diseases that can affect both muscles and the nervous system, such as inflammatory myopathies, autoimmune diseases, and metabolic myopathies including mitochondrial disorders. The presence of ragged red fibers on muscle biopsy suggestive of mitochondrial myopathy. I will firstly lookup diseases I suspect. \textcolor{reasoncolor}{\texttt{</reason>}}

\vspace{0.3cm}
\textcolor{lookupcolor}{\texttt{<lookup>}} inflammatory myopathy, autoimmune disease, mitochondrial myopathy... \textcolor{lookupcolor}{\texttt{</lookup>}}

\vspace{0.2cm}
\texttt{<guide>}
**inflammatory myopathy**: proximal muscle weakness, elevated muscle enzymes (CK, aldolase), muscle pain and tenderness, characteristic rashes in dermatomyositis, dysphagia, and interstitial lung disease. Major types include dermatomyositis, polymyositis, and inclusion body myositis...

**autoimmune disease**: fatigue, fever, joint pain, organ-specific manifestations, positive autoantibodies (ANA, anti-dsDNA, RF), and multi-system involvement affecting skin, joints, kidneys, and nervous system...

**mitochondrial myopathy**: progressive muscle weakness, exercise intolerance, ptosis, external ophthalmoplegia, proximal muscle weakness, and characteristic ragged red fibers on muscle biopsy...

...
\texttt{</guide>}

\vspace{0.3cm}
\textcolor{reasoncolor}{\texttt{<reason>}} The search results suggest three main categories to consider. Inflammatory myopathies could explain the proximal muscle weakness, muscle atrophy, and elevated CSF protein through inflammatory muscle disease with potential CNS involvement. Autoimmune diseases can cause multi-system manifestations including neurologic symptoms, proteinuria, and muscle weakness through immune-mediated tissue damage. Mitochondrial myopathy is particularly compelling given the presence of ragged red fibers, ptosis, and progressive muscle weakness, though the inflammatory features and systemic involvement suggest a mixed or overlapping pathology... \textcolor{reasoncolor}{\texttt{</reason>}}

\vspace{0.3cm}
\textcolor{matchcolor}{\texttt{<match>}} ptosis, numbness and tingling, fatigue, proximal muscle weakness, muscle weakness, numbness, tingling, fatigue, abnormal erythrocyte sedimentation rate, elevated protein in cerebrospinal fluid, muscle atrophy, decreased reflexes, distal sensory impairment \textcolor{matchcolor}{\texttt{</match>}}

\vspace{0.2cm}
\texttt{<refer>}
Phenotype: Bilateral ptosis, proximal muscle weakness, limb-girdle muscle weakness, decreased tendon reflexes, reduced lower limb muscle strength; tingling sensation, distal upper limb weakness, elevated cerebrospinal fluid protein. Confirmed disease: [Guillain-Barré syndrome]

Phenotype: Proximal muscle weakness, distal muscle weakness, peripheral sensorimotor neuropathy, elevated cerebrospinal fluid protein, pain, numbness, tingling, progressive motor decline. Confirmed disease: [Chronic inflammatory demyelinating polyneuropathy]

Phenotype: Paresthesia, muscular weakness, foot drop, distal lower limb weakness, reduced sensation, areflexia, elevated erythrocyte sedimentation rate, elevated C-reactive protein level. Confirmed disease: [Eosinophilic Granulomatosis with Polyangiitis]

Phenotype: Polymyositis, muscle weakness, cardiac involvement in inflammatory myopathy, interstitial lung disease, elevation of serum creatine phosphokinase. Confirmed disease: [Polymyositis]

...
\texttt{</refer>}

\vspace{0.3cm}
\textcolor{reasoncolor}{\texttt{<reason>}} The phenotypes in the reference cases suggest several diseases that could fit the patient's symptoms. Guillain-Barré syndrome and chronic inflammatory demyelinating polyneuropathy could explain the neuropathic features and CSF abnormalities. Eosinophilic granulomatosis with polyangiitis can cause peripheral neuropathy and muscle weakness with systemic inflammation... However, some typical patterns can not be well-explained and worth further retrieval...\textcolor{reasoncolor}{\texttt{</reason>}}

\vspace{0.3cm}
\textcolor{matchcolor}{\texttt{<match>}} muscle weakness, numbness, tingling, fatigue, proteinuria, elevated protein in cerebrospinal fluid, decreased reflexes, distal sensory impairment, inflammatory myopathy, eosinophilic granulomatosis with polyangiitis \textcolor{matchcolor}{\texttt{</match>}}

\vspace{0.2cm}
\texttt{<refer>}
Phenotype: Tingling sensation, distal upper limb weakness, normal reflexes, normal sensory examination, elevated cerebrospinal fluid protein. Confirmed disease: [Acute Inflammatory Demyelinating Polyneuropathy]

Phenotype: Proximal muscle weakness, distal muscle weakness, peripheral sensorimotor neuropathy, elevated cerebrospinal fluid protein, oligoclonal bands, pain, numbness, tingling. Confirmed disease: [Chronic inflammatory demyelinating polyneuropathy]

Phenotype: Fever, myalgia, weight loss, fatigue, anorexia, arthralgia, anemia, elevated C-reactive protein, proteinuria, inflammatory myopathy, polyserositis. Confirmed disease: [Systemic Lupus Erythematosus]

...
\texttt{</refer>}

\vspace{0.3cm}
\textcolor{reasoncolor}{\texttt{<reason>}} The matching cases from the second search further refine our differential diagnosis. The conditions consistently appearing include inflammatory demyelinating neuropathies, systemic autoimmune diseases with muscle involvement, and endocrine-related myopathies. The combination of peripheral neuropathy, muscle weakness, and systemic inflammation points toward autoimmune etiologies. Let's search for more specific information about the relationship between these conditions and the patient's key symptoms. \textcolor{reasoncolor}{\texttt{</reason>}}

\vspace{0.3cm}
\textcolor{lookupcolor}{\texttt{<lookup>}} dermatomyositis, systemic lupus erythematosus, ... \textcolor{lookupcolor}{\texttt{</lookup>}}

\vspace{0.2cm}
\texttt{<guide>}
**dermatomyositis**: Characterized by proximal muscle weakness, characteristic skin rashes (heliotrope rash, Gottron's papules), elevated muscle enzymes, dysphagia, and potential for malignancy association...

**systemic lupus erythematosus**: SLE can present with inflammatory myopathy in 5-10\% of cases, featuring muscle weakness, elevated CK, myalgia, along with other SLE manifestations like proteinuria, neurologic involvement, and positive autoantibodies...

...
\texttt{</guide>}

\vspace{0.3cm}
\textcolor{searchcolor}{\texttt{<search>}} |PubMed| relationship between ptosis and inflammatory myopathies, |PubMed| SLE neurologic manifestations with ptosis, ... \textcolor{searchcolor}{\texttt{</search>}}

\vspace{0.2cm}
\texttt{<result>} Ptosis in inflammatory myopathies can occur through several mechanisms: direct extraocular muscle involvement in dermatomyositis/polymyositis, cranial nerve involvement in systemic vasculitis, or neuromuscular junction dysfunction. SLE can cause ptosis through cranial neuropathy, CNS lupus affecting brainstem nuclei, or concurrent myasthenia gravis (increased association).

...
\texttt{</result>}

\vspace{0.3cm}
\textcolor{diagnosecolor}{\texttt{<diagnose>}} The following diseases should be of primary concern and warrant further investigation: \textbf{Dermatomyositis}, \textbf{Systemic Lupus Erythematosus}, \textbf{Mitochondrial Myopathy}... \textcolor{diagnosecolor}{\texttt{</diagnose>}}
\end{tcolorbox}

\subsection{Details in Data Processing per Resource}
\label{supp:data_resource}

\begin{figure}[t]
    \centering
    \includegraphics[width=1\linewidth]{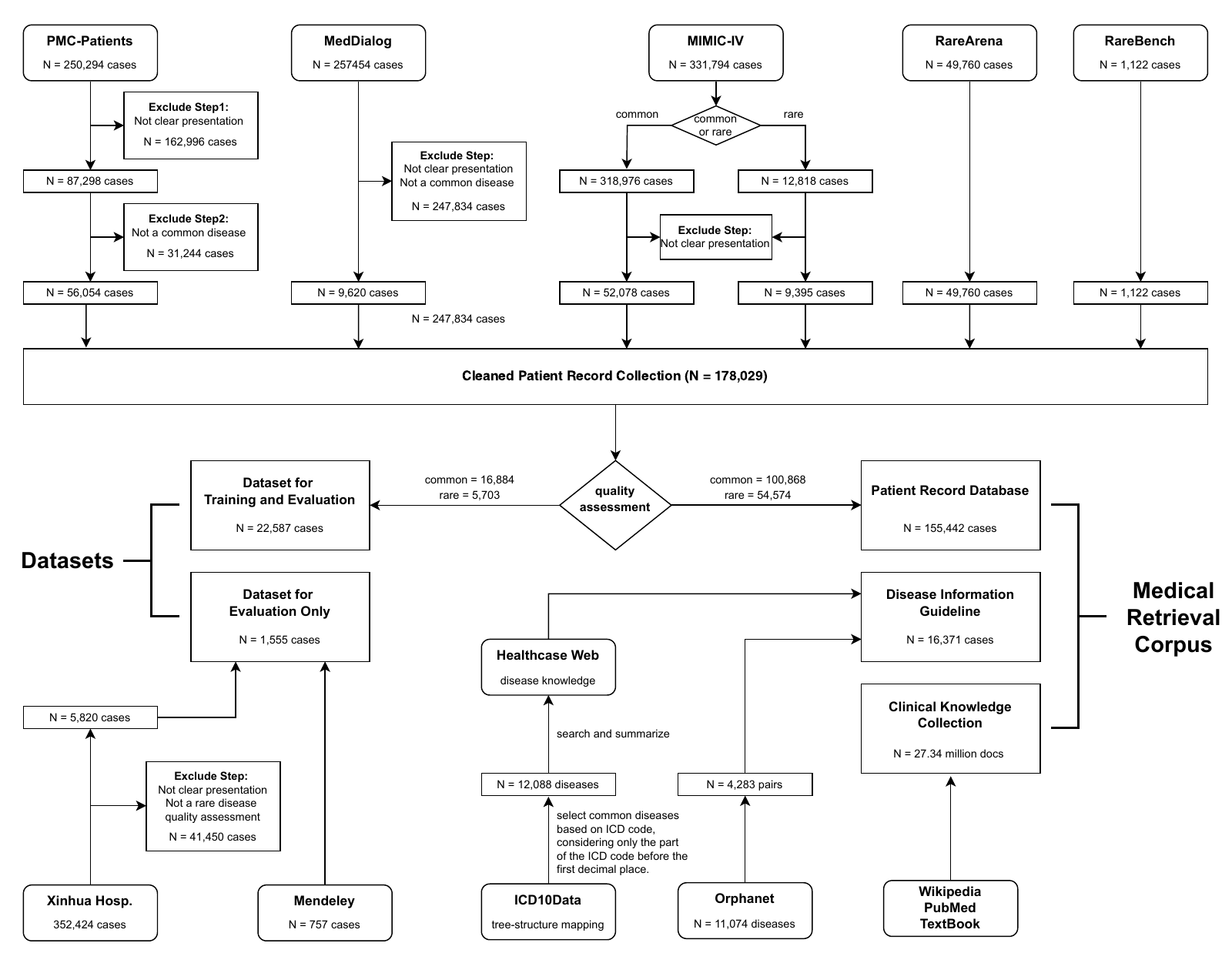}
    \caption{\textbf{Data processing procedure.}
    The datasets for training and evaluation are derived from eight data sources and are split into training, evaluation, and evaluation-only sets. The medical retrieval corpus is constructed partially from these datasets as well as additional authoritative online resources.}
    \label{fig:dataprocess_supp}
\end{figure}

We utilized a diverse set of clinical and biomedical resources to construct the training and evaluation datasets, as well as the retrieval corpus~(Supplementary Fig.~\ref{fig:dataprocess_supp}). Below, we describe each source in terms of its origin, processing steps, and specific use within our framework, with their license presented in Supplementary Tab.~\ref{tab:data_compliance_supp}.

\textbf{MIMIC-IV}~\cite{Johnson2023MIMICIVAF}. This public dataset contains 331,794 de-identified discharge summaries from 145,915 patients admitted to Beth Israel Deaconess Medical Center. We first categorize these cases into common and rare diseases based on ICD codes and primary diagnoses, following the disease classification Stage~(Stage 1 in Sec.~\ref{sec:data_description}). This process yields 318,976 common and 12,818 rare disease cases. 
To ensure data integrity, we utilize GPT-4o to evaluate case quality, excluding entries with ``low-quality'' attributes, such as those lacking a clear causal link between the clinical presentation and the final diagnosis~(Stage 2 in Sec.~\ref{sec:data_description}). After this filtering, 52,078 common and 9,395 rare disease cases remain. Following the standards of extraction and stratification~(Stage 3 in Sec.~\ref{sec:data_description})in, we identify 7,257 cases for the common disease diagnosis task and 2,184 for the rare disease task. The remaining 44,821 common and 7,211 rare cases are integrated into the patient record database used by \ModelName{}.

\textbf{PMC-Patients}~\cite{Zhao2022PMCPatientsAL}. This dataset comprises 250,294 patient profiles derived from 167,000 public summaries in PubMed Central. We first employ GPT-4o to assess the quality of these cases, excluding 162,995 entries identified as ``low-quality.''~(Stage 2). Of the remaining 87,298 cases, we categorize them~(Stage 1) and retain only those associated with common diseases, resulting in 56,054 cases. Following our extraction and stratification pipeline~(Stage 3), we select 6,421 cases for the common disease diagnosis task, while the remaining 49,633 cases are incorporated into the patient record database for \ModelName{}.

\textbf{MedDialog}~\cite{Chen2020MedDialogAL}. This dataset contains clinical consultations from both Chinese- and English-speaking online platforms. We utilize the English subset, which initially comprises 257,454 cases. By applying the aforementioned pipeline to exclude rare diseases~(Stage 1) and identify ``low-quality'' entries~(Stage 2), we remove 247,839 cases, leaving 9,620 cases associated with common diseases. Following the diagnostic data processing pipeline~(Stage 3), 3,206 of these cases are designated for the training and testing tasks. The remaining 6,414 cases are integrated into the patient record database for \ModelName{}.

\textbf{RareArena}~\footnote{\url{https://github.com/zhao-zy15/RareArena}}. Derived from PMC-Patients, this dataset contains approximately 50,000 patient records covering over 4,000 diseases. As these cases are pre-curated for rare disease tasks, we bypass the processing stage 1 \& 2. However, we apply the extraction and stratification pipeline, which explicitly identifies 3,242 cases for training and evaluation. The remaining 46,518 cases are incorporated into the patient record database.

\textbf{RareBench}~\cite{Chen2024RareBenchCL}. This benchmark targets rare disease diagnosis and includes both public and private components. We utilize 1,122 cases from the public sources (RAMEDIS, MME, HMS, and LIRICAL). Given this dataset provides structured phenotype-diagnosis fields and explicitly targets rare diseases, we do not apply the additional quality assessment or filtering pipeline. From this collection, we randomly select 798 cases for training and evaluation, while the remaining 324 cases are integrated into the patient record database.

\textbf{Mendeley}~\cite{Shafi2025ASD}. Released in June 2025, this structured resource details binary associations between 85 common diseases and 172 symptoms derived from peer-reviewed literature and reputable databases. We utilize this dataset for zero-shot evaluation; its release date postdates the training cutoffs of the models, ensuring no prior exposure and allowing for an assessment of generalization to new data. Unlike the previously described datasets, we bypass the additional GPT-4o quality assessment here, as the highly structured curation of this resource inherently minimizes noise.

\textbf{Xinhua Hosp}~\cite{Zhao2025AnAS}. This in-house dataset comprises rare disease diagnostic records from \textit{Xinhua Hospital Affiliated To Shanghai Jiao Tong University School of Medicine} spanning 2014 to 2025, totaling 352,424 entries. We apply the GPT-4o quality assessment to exclude ``low-quality'' entries~(Stage 2), followed by the extraction and stratification pipeline~(Stage 3), which yields a total of 5,820 validated cases. From this subset, we randomly sample 798 cases for evaluation. Distinct from the public datasets, the remaining cases are not integrated into the patient record database during training; instead, they function exclusively as an extra retrieval source during zero-shot testing.


\textbf{ICD10Data}. 
We extracted disease names and codes from the official ICD-10-CM classification, yielding 12,088 common and 4,283 rare diseases. This taxonomy was used to construct our disease information guide.

\textbf{Orphanet}. 
We obtained 11,074 Orpha codes, including phenotype probability distributions for 4,283 rare diseases. These were integrated into the structured knowledge base to support phenotype-driven reasoning.

\textbf{Healthcare Websites}. 
We curated disease descriptions, symptoms, and other clinical features from online medical sources~({\em e.g.}, NCBI, WebMD, NIH, Mayo Clinic). 
Using deepseek-v3, we summarized and standardized 142,141 disease–symptoms/phenotypes pairs for inclusion in the structured guideline.

\textbf{PubMed, Wikipedia, and Textbooks}.
Following the MedRAG protocol~\cite{xiong2024benchmarking}, we aggregated 23.9 million PubMed abstracts and 3.31 million Wikipedia medical entries and 18 medical textbooks to form a broad clinical knowledge base. They were further chunked and indexed into database to facilitate efficient retrieval. 

\begin{table}[h]
\centering
\scriptsize
\caption{Legal audit of training \& evaluation datasets.}
\label{tab:data_compliance_supp}
\renewcommand{\arraystretch}{1.3}
\begin{tabular}{p{0.2\linewidth} p{0.25\linewidth} p{0.45\linewidth}}
\toprule
\textbf{Dataset} & \textbf{License Type} & \textbf{Permission for LLM Training} \\
\midrule
\textbf{MIMIC-IV} & PhysioNet DUA v1.5.0 (Credentialed) & \textbf{Authorized.} DUA permits derived models; compliant with PhysioNet LLM guidelines. \\
\midrule
\textbf{PMC-Patients} & CC BY / CC BY-SA & \textbf{Authorized.} Part of the Open Access Subset specifically for text mining. \\
\midrule
\textbf{RareBench} & Apache 2.0 & \textbf{Authorized.} Permissive license allowing modification and derivative works. \\
\midrule
\textbf{RareArena} & CC BY-NC-SA 4.0 & \textbf{Authorized.} Permits non-commercial derivative works. \\
\midrule
\textbf{Mendeley (Bangla)} & CC BY 4.0 & \textbf{Authorized.} Permits unrestricted use with attribution. \\
\midrule
\textbf{MedDialog} & Academic Research Use & \textbf{Fair Use.} Used strictly for non-commercial academic validation. \\
\bottomrule
\end{tabular}
\end{table}

\subsection{Details in Retrieval Methods}
To maximize the efficacy and performance of the interaction with our proposed medical retrieval corpus, we treat each retrieval action and the observation of the action as tool and input arguments. Here we detaied the formulation of these tools including the Phenotype Parser, Patient Matcher, knowledge Searcher and MedDoc Summarizer.

\noindent \textbf{Phenotype Parser.} This tool is designed for the retriving from the diseae information guideline.
We use BM25 search algorithm to build this tool for phenotype parsing with the input of a list of diseases. To optimize the response time, we process it batch by batch for searching process acceleration. Specifically, take $\mathcal{D} = \{d_1, d_2, ..., d_m\}$ as input where $d_i$ denotes the $i^{th}$ disease waiting for searching, then the general process could be denoted as:

\begin{align}
T_{\mathrm{PP}}(\mathcal{D}) = 
\Bigg\{
\left(
d,\, 
\begin{cases}
    \mathcal{P}(\hat{d}), & \text{if } \mathrm{BM25}(d, \hat{d}) \geq \tau \\
    \text{no reference}, & \text{otherwise}
\end{cases}
\right)
\,\Bigg|\, 
d \in \mathcal{D},\ \hat{d} = \underset{d' \in \mathcal{M}_{\mathrm{disease}}}{\arg\max}\; \mathrm{BM25}(d, d')
\Bigg\}
\end{align}

Here, $\operatorname{BM25\_Match}(d, \mathcal{M}_{\mathrm{disease}})$ denotes the best-matching disease $\hat{d}$ for a query $d$ in the reference corpus $\mathcal{M}_{\mathrm{disease}}$ using the standard BM25 algorithm, where the BM25 score between a tokenized query $q$ and a candidate disease name $d'$ is defined as 
\[
\mathrm{BM25}(q, d') = \sum_{t \in q} \mathrm{IDF}(t) \cdot \frac{f(t, d')\,(k_1 + 1)}{f(t, d') + k_1(1 - b + b \frac{|d'|}{\mathrm{avgdl}})}
\]
with $f(t, d')$ being the frequency of token $t$ in $d'$, $|d'|$ the number of tokens in $d'$, $\mathrm{avgdl}$ the average length of all disease names in the corpus, and $k_1$, $b$ standard hyperparameters (e.g., $k_1 = 1.5$, $b = 0.75$). The inverse document frequency is computed as 
\[
\mathrm{IDF}(t) = \log \left( \frac{N - n(t) + 0.5}{n(t) + 0.5} + 1 \right),
\]
where $N$ is the total number of diseases and $n(t)$ is the number of diseases containing token $t$. For each $d \in \mathcal{D}$, if the maximum BM25 score $\mathrm{BM25}(d, \hat{d})$ exceeds a threshold $\tau$, we return the top $k$ (e.g., $k=10$) high-frequency phenotypes for the matched disease, denoted as $\mathcal{P}(\hat{d})$; otherwise, we return ``no reference''.

\noindent \textbf{Patient Matcher.} This tool is designed to interact with the patient record database When taking symptoms or phenotypes as input, matching to patients in similar situations can provide valuable references for current case diagnosis. Given that different patients may describe symptoms differently, lexical searching is not adopted. Instead, we use BioLORD embeddings to calculate semantic similarity between cases. Specifically, each phenotype or symptom $s$ in a patient record is encoded as a feature vector $\mathbf{e}(s)$ using the BioLORD encoder. For a case $i$ with set $\mathcal{P}_i = \{p_{i,1}, p_{i,2}, \ldots, p_{i,n_i}\}$, we represent its overall case embedding as the transformation of the symptom embeddings:

\begin{align}
    \mathrm{Sim}(\mathcal{P}_q, \mathcal{P}_i) = 
    \frac{1}{|\mathcal{P}_q|} \sum_{j=1}^{|\mathcal{P}_q|} \max_{1 \leq k \leq |\mathcal{P}_i|} 
    \cos\left( \mathbf{e}(p_{q,j}), \mathbf{e}(p_{i,k}) \right)
\end{align}

where $\mathcal{P}_q = \{p_{q,1}, \ldots, p_{q,n_q}\}$ is the query case, $\mathcal{P}_i = \{p_{i,1}, \ldots, p_{i,n_i}\}$ is the $i$-th case in the database, and $\cos(\mathbf{a}, \mathbf{b})$ denotes the cosine similarity between two embedding vectors. For each query symptom $p_{q,j}$, we find its maximal similarity to all symptoms in the candidate case, and then average these maxima across all query symptoms.

The Patient Matcher tool $T_\mathrm{PM}$ returns the top-$N$ cases with the highest similarity scores:
\begin{align}
    T_\mathrm{PM}(\mathcal{P}_q) = \operatorname{TopN}_i\left( \mathrm{Sim}(\mathcal{P}_q, \mathcal{P}_i) \right)
\end{align}

where $\operatorname{TopN}_i(\cdot)$ selects the $N$ most similar cases from the database.

\noindent \textbf{Knowledge Searcher.} This tool is designed to interact with medical knowledge collection.
To deploy these corpora as an efficient retrieval service accessible to Large Language Models (LLMs), we developed an asynchronous web server using the Python-based FastAPI framework, served by Uvicorn. The system implements two mainstream retrieval paradigms: sparse retrieval, based on keyword frequency (BM25), and dense retrieval, based on semantic similarity. For sparse retrieval, we leveraged the Pyserini library to query a pre-constructed Lucene index. For dense retrieval, we first utilized the Transformers library to load pre-trained text embedding models (e.g., E5, BGE) to encode all text chunks into high-dimensional vectors. Subsequently, we employed the FAISS library to build an index for these vectors, enabling millisecond-level similarity searches across a massive vector space by leveraging its GPU acceleration capabilities. The server's core logic is encapsulated within an Encoder class and multiple Retriever classes; the former handles text vectorization, while the latter executes the specific retrieval operations (either BM25 or Dense) based on the provided configuration. It calls a batch\_search method to perform real-time query encoding and retrieves the top-k most relevant documents from the corresponding FAISS or Lucene index, returning the final results in JSON format. The entire service is initiated via a command-line script, which allows for flexible configuration of key parameters such as index and corpus paths, the choice of retrieval model, and the number of documents to return (top-k). This design results in a highly configurable, scalable, and high-performance retrieval backend.

\noindent \textbf{MedDoc Summarizer.} 
To mitigate context length constraints when environment feedback exceeds maximum limits, we employed a document summarizer to condense information into a controllable length.
Specifically, we utilized \textbf{Qwen-3-235B} as the summarizer, deployed within our multi-source environment using the \texttt{SGLang} framework for efficient batch inference.
This inference infrastructure was hosted on four NVIDIA H200 GPUs within the \ModelName{} training cluster.
The summarization instruction prompt is provided below:

\begin{tcolorbox}[colback=gray!5!white, colframe=blue!50!black, width=0.98\textwidth, sharp corners=south]
\textbf{System Prompt:}

You are a medical document summarization assistant. Given a search query and a retrieved document, your task is to summarize the document to directly and concisely answer the query.

\begin{itemize}
    \item Extract the most relevant facts or statements from the document that directly answer the query. If more than 10 points are relevant, keep only the 10 most important.
    \item Your answer should be brief, focused, and contain no extra explanation.
    \item Format your answer as a JSON string, e.g., \texttt{"answer": "..."}.
    \item If no relevant information can be found, respond with \texttt{"answer": "no reference"}.
\end{itemize}

\textbf{Agent:} \\
\texttt{Source: \{source\} \ \textbar\  Query: \{query\}}
\end{tcolorbox}

\subsection{Prompt for Reasoning Quality Assessment}
\label{supp:reasoning_assess_prompt}
To quantitatively evaluate the diagnostic capabilities, we employed advanced LLMs, specifically DeepSeek-R1 and Meditron, as evaluator agents. The prompt presented below was designed to instruct these evaluators to assess the entire diagnostic reasoning process of the target models for a given patient case, strictly adhering to the proposed five-dimensional scoring framework.

\begin{mdframed}[
    linewidth=2pt,
    linecolor=black,
    backgroundcolor=gray!5,
    skipabove=10pt,
    skipbelow=10pt,
    innertopmargin=15pt,
    innerbottommargin=15pt,
    innerleftmargin=15pt,
    innerrightmargin=15pt
]

\textbf{Role \& Objective:}
You are an expert Medical Evaluator. Your task is to evaluate the clinical reasoning capabilities of an AI model based on a specific patient case and a ground truth diagnosis. You must strictly adhere to the following 5-dimensional framework.

\vspace{0.5em}
\textbf{Evaluation Criteria (Scale 1-4):}

\begin{enumerate}[leftmargin=*, label=\textbf{\arabic*.}, itemsep=4pt]

    \item \textbf{Overall Correctness (Diagnostic Accuracy)}
    \begin{itemize}[leftmargin=1.5em, label={\scriptsize$\bullet$}, nosep]
        \item \textbf{Score 1 (Poor):} The diagnosis is incorrect, or the model provides dangerous/contraindicated recommendations.
        \item \textbf{Score 2 (Fair):} Identifies the general disease category or system but misses the specific pathology; includes unreasonable differentials.
        \item \textbf{Score 3 (Good):} Correct diagnosis is present but ranked low in the candidate list, or the list contains significant distractors.
        \item \textbf{Score 4 (Excellent):} Precisely identifies the ground truth as the primary diagnosis and successfully excludes reasonable differentials.
    \end{itemize}

    \item \textbf{Clinical Utility (Actionability)}
    \begin{itemize}[leftmargin=1.5em, label={\scriptsize$\bullet$}, nosep]
        \item \textbf{Score 1 (Poor):} Response lacks actionable insight, is disorganized, or increases the physician's cognitive load.
        \item \textbf{Score 2 (Fair):} Provides generic, ``textbook-style'' information not specifically tailored to the patient's unique case presentation.
        \item \textbf{Score 3 (Good):} Integrates case-specific details, providing valuable reference points that aid the diagnostic process.
        \item \textbf{Score 4 (Excellent):} Achieves a ``Consultant'' level of insight, offering the correct diagnosis alongside profound analysis and management recommendations.
    \end{itemize}

    \item \textbf{Reasoning Consistency (Logical Flow)}
    \begin{itemize}[leftmargin=1.5em, label={\scriptsize$\bullet$}, nosep]
        \item \textbf{Score 1 (Poor):} Logic is chaotic, exhibiting non-sequiturs or significant logical fallacies.
        \item \textbf{Score 2 (Fair):} Relies on superficial keyword matching without comprehensive clinical synthesis or analysis.
        \item \textbf{Score 3 (Good):} Logic is clear, establishing reasonable causal links between presenting symptoms and potential diseases.
        \item \textbf{Score 4 (Excellent):} Demonstrates complex clinical reasoning (e.g., diagnosis by exclusion, multi-system synthesis) with robust logical coherence.
    \end{itemize}

    \item \textbf{Reference Info Relevance (Signal-to-Noise Ratio)}
    \begin{itemize}[leftmargin=1.5em, label={\scriptsize$\bullet$}, nosep]
        \item \textbf{Score 1 (Poor):} Discussion is severely off-topic, focusing on irrelevant organ systems or non-medical content.
        \item \textbf{Score 2 (Fair):} Output contains excessive irrelevant background knowledge or generic, non-substantive text.
        \item \textbf{Score 3 (Good):} Content is primarily focused on the specific case with minimal redundancy.
        \item \textbf{Score 4 (Excellent):} Highly pertinent; every step is tightly aligned with the diagnostic needs of the current case (High Signal-to-Noise Ratio).
    \end{itemize}

    \item \textbf{Hallucination Severity (Faithfulness)}
    \begin{itemize}[leftmargin=1.5em, label={\scriptsize$\bullet$}, nosep]
        \item \textbf{Score 1 (Severe):} Fabricates symptoms, physical signs, or laboratory results not present in the patient record (Fatal Errors).
        \item \textbf{Score 2 (Moderate):} Confuses medical concepts or misattributes information (e.g., citing incorrect guidelines).
        \item \textbf{Score 3 (Minor):} Slight deviations in details (e.g., severity, frequency) that do not fundamentally alter the diagnostic direction.
        \item \textbf{Score 4 (None):} Completely faithful to the case description; all cited medical knowledge and patient data are accurate.
    \end{itemize}

\end{enumerate}

\vspace{0.5em}
\hrule
\vspace{0.5em}

\textbf{Output Requirements:}
\begin{itemize}[leftmargin=*, nosep]
    \item You must output the result strictly as a \textbf{JSON object}. Do not include any conversational text.
    \item The JSON object must contain the following keys: \texttt{overall\_correctness}, \texttt{clinical\_utility}, \texttt{reasoning\_consistency}, \texttt{reference\_relevance}, \texttt{hallucination\_severity}.
    \item Each key must contain a \texttt{"score"} (integer 1-4) and a \texttt{"justification"} (concise explanation referencing the criteria).
\end{itemize}

\vspace{0.5em}
\textbf{JSON Example:}

\begin{Verbatim}[frame=single, fontsize=\small]
{
  "overall_correctness": {
      "score": 3,
      "justification": "Correct diagnosis is present but ranked low..."
  },
  "clinical_utility": {
      "score": 2,
      "justification": "Provides generic info not tailored to the patient..."
  },
  ...
}
\end{Verbatim}

\end{mdframed}

\subsection{Top-N Accuracy Calculation Protocol}
\label{supp:eval_protocol}

Given the inherent variability of natural language generation, 
establishing a rigorous performance metric is essential for reproducible clinical evaluation. We implemented a standardized \textbf{two-stage protocol} comprising deterministic extraction followed by rule-enhanced strict matching. This approach ensures that reported Top-N accuracy metrics reflect genuine diagnostic capability, decoupling clinical reasoning performance from formatting variability or semantic ambiguity.

\subsubsection{Deterministic Extraction via Format Enforcement}

In contrast to standard generative settings where answers may be dispersed within unstructured conversational text, our framework imposes a rigorous structural constraint reinforced during training via the format reward ($\sigma_{f}$). The model is required to encapsulate the final diagnostic conclusion within \texttt{<diagnose>} tags and explicitly delineate specific disease entities using \LaTeX{}-style bold formatting (\texttt{\textbackslash textbf\{\}}). For instance, a valid output must strictly adhere to the schema: \texttt{<diagnose> ... \textbackslash textbf\{Disease A\} ... </diagnose>}. We parse the content enclosed within these bold delimiters to generate an ordered candidate list $C = [c_1, c_2, \dots, c_k]$, treating any text outside these tags as auxiliary reasoning context excluded from metric calculation. Adherence is strictly enforced; any response failing to satisfy these formatting constraints is assigned a score of zero for that instance.

\subsubsection{Rule-Enhanced Strict Matching}

To evaluate the correspondence between a candidate diagnosis $c_i$ and the ground-truth $g$, we employ a strict matching protocol that rejects vague semantic similarities while accommodating clinically valid specificity refinements. The process begins with standard string normalization for both the candidate and ground-truth, including case-folding and the removal of punctuation and whitespace. 

\textbf{Permissible clinical variations.} 
We acknowledge that a candidate diagnosis may be clinically correct even if it includes specific modifiers absent in a more generic ground-truth. Therefore, a candidate $c_i$ is considered a match if it consists of the ground-truth string $g$ modified exclusively by terms from a predefined set of permissible clinical variations. These variations span five axes: \textit{chronicity and course} ({\em e.g.}, acute, recurrent, progressive); 
\textit{severity and state} ({\em e.g.}, mild, malignant, latent); 
\textit{etiology and origin} ({\em e.g.}, idiopathic, congenital, iatrogenic); \textit{laterality and localization} ({\em e.g.}, bilateral, focal, systemic); and \textit{typing and staging} ({\em e.g.}, Type I, Stage 4). Formally, if $g$ is a substring of $c_i$, and the set difference consisting of the remaining words belongs solely to these allowable categories, the prediction is counted as correct.

\textbf{Exclusion of fuzzy matches.} To maintain the highest evaluation standards, we enforce exact character matching for the core disease entity, deliberately excluding probabilistic semantic metrics such as BERTScore or edit-distance measures like Levenshtein distance. 
This strict protocol penalizes orthographic variations ({\em e.g.}, ``\textit{Haemophilia}'' vs.~``\textit{Hemophilia}'') and synonymous terminology ({\em e.g.}, ``Hepatic Steatosis'' vs.~``Fatty Liver'') unless explicitly linked within the source taxonomy. 
While our training objective utilizes partial rewards to facilitate optimization, 
the testing protocol remains binary and stringent, ensuring that high Top-N accuracy reflects precise adherence to standardized medical terminology.

\subsection{System prompt for baseline models}

For vanilla model with direct diagnosis inference, we use the following prompt:

\begin{tcolorbox}[colback=gray!5!white, colframe=black, width=0.98\textwidth, sharp corners=south]
\textbf{System Prompt:}

You are a disease diagnosis assistant. Your task is to make diagnosis based on the given symtoms or phenotypes. The user input is a list of symptoms of phenotypes. Your answer should only be diseases without other explanations enclosed within LaTeX bold format: \textbf{Disease1}, \textbf{Disease2}, etc. Please make up to 5 diagnoses.

\end{tcolorbox}

For training-free RAG, we use the same prompt as detailed in Supp.~\ref{inital_prompt}.

\end{document}